\documentclass[lettersize,journal]{IEEEtran}
\usepackage{amsmath,amsfonts}
\usepackage{algorithmic}
\usepackage{algorithm}
\usepackage{array}
\usepackage[caption=false,font=normalsize,labelfont=sf,textfont=sf]{subfig}
\usepackage{textcomp}
\usepackage{stfloats}
\usepackage{graphicx}
\usepackage{tabularx}
\usepackage{booktabs}
\usepackage{url}
\usepackage{multirow}
\usepackage{verbatim}

\usepackage{cite}
\begin{document}

\title{Ultra-short-term solar power forecasting by deep learning and data reconstruction}

\author{IEEE Publication Technology,~\IEEEmembership{Staff,~IEEE,}
\thanks{This paper was produced by the IEEE Publication Technology Group. They are in Piscataway, NJ.}
\thanks{Manuscript received April 19, 2021; revised August 16, 2021.}}

\author{Jinbao Wang, Jun Liu, Shiliang Zhang, Xuehui Ma
\thanks{J. Wang, J. Liu and X. Ma are with Xi’an University of Technology, Xi'an, China (email: jbwang@stu.xaut.edu.cn, liujun0310@xaut.edu.cn, xuehui.yx@gmail.com), S. Zhang is with University of Oslo, Norway (email: shilianz@uio.no)}
}


\maketitle

\begin{abstract}
The integration of solar power has been increasing as the green energy transition rolls out. The penetration of solar power challenges the grid stability and energy scheduling, due to its intermittent energy generation. Accurate and near real-time solar power prediction is of critical importance to tolerant and support the permeation of distributed and volatile solar power production in the energy system. 
In this paper, we propose a deep-learning based ultra-short-term solar power prediction with data reconstruction. We decompose the data for the prediction to facilitate extensive exploration of the spatial and temporal dependencies within the data. Particularly, we reconstruct the data into low- and high-frequency components, using ensemble empirical model decomposition with adaptive noise (CEEMDAN). We integrate meteorological data with those two components, and employ deep-learning models to capture long- and short-term dependencies towards the target prediction period. In this way, we excessively exploit the features in historical data in predicting a ultra-short-term solar power production. 
Furthermore, as ultra-short-term prediction is vulnerable to local optima, we modify the optimization in our deep-learning training by penalizing long prediction intervals. 
Numerical experiments with diverse settings demonstrate that, compared to baseline models, the proposed method achieves improved generalization in data reconstruction and higher prediction accuracy for ultra-short-term solar power production. 
\end{abstract}


\section{Introduction}
\IEEEPARstart{I}{n} In order to achieve sustainable development goals, countries around the world are undergoing an energy transition to reduce dependence on fossil fuels and greenhouse gas emissions. PV power generation has become an important alternative to fossil fuels due to its clean, renewable nature and wide distribution \cite{zhang2023evaluation,wang2024integrated,duguma2023privacy}. More than 75\% of the global new renewable energy capacity in 2023 came from PV power generation, with approximately 60\% of the new renewable energy generation originating from PV systems. Among these, rooftop solar installations can maximize the utilization of solar energy resources, characterized by low access barriers and high flexibility \cite{zhai2024scheduling,zhai2025photovoltaic}. Though more energy datasets are available in facilitating solar power analyses \cite{zhang2025norwegi} and enhance energy efficiency, however, the intermittent and highly volatile characteristics of PV power generation introduce substantial uncertainty to grid scheduling and energy management \cite{liu2023evolution,gong2025ensemble,10738058}. Therefore, in the context of the integration of the large-scale PV grid, there is an urgent need for accurate and widely applicable PV power prediction technology to ensure normal grid operation and reduce energy losses \cite{gong2025parallel}.

For PV output power prediction, it can be categorized into three types from a temporal scale perspective: long-term, medium-term, short-term, and ultra-short-term prediction \cite{song2024graph,gong2025parallel}. Ultra-short-term prediction typically covers periods from several minutes to several hours, requiring the highest prediction accuracy and presenting the greatest technical challenges \cite{zhai2025photovoltaic}. It plays a crucial role in instantaneous grid scheduling, short-term electricity market trading, and real-time power balance scenarios \cite{li2023ultra,zhang2025data}. From the perspective of prediction results, it can be divided into two types: deterministic prediction and probabilistic prediction. Deterministic prediction can only provide precise point prediction information, while probabilistic prediction can cover an uncertainty range \cite{10551447} of future PV power generation and occurrence probabilities of various output levels, offering decision-makers more information to make better decisions \cite{arora2022probabilistic,song2025probabilistic,tawn2022review}. This research focuses on ultra-short-term PV power generation probabilistic prediction.

Currently, the main prediction models for ultra-short-term PV probabilistic forecasting are ensemble probabilistic models, statistical probabilistic models, and deep learning probabilistic models. Traditional statistical models mainly include quantile regression \cite{bozorg2020bayesian,ma2022adaptive}, random forest \cite{ibrahim2019optimized}, and Bayesian methods \cite{doubleday2020probabilistic,ma2022active}. These models have simple structures and strong interpretability, but they have difficulty simultaneously handling multi-scale temporal features of PV data, resulting in relatively low prediction accuracy. Therefore, deep learning probabilistic models have received widespread attention \cite{husein2024towards}.

Deep learning models include convolutional neural networks \cite{hong2023short}, long short-term memory (LSTM) networks \cite{da2023comparing,10901967}, transformer models \cite{tao2024operational}, and gated recurrent units (GRU) \cite{zarzycki2022advanced}. These methods possess powerful feature extraction capabilities and have demonstrated excellent performance in handling complex prediction tasks and large-scale datasets. For example, transformer models have shown outstanding performance in medium-to-long-term prediction due to their self-attention mechanism and streamlined architecture \cite{liu2025novel}. Research has shown that combining the output features of multiple models can further improve the model's prediction performance \cite{yu2024adaptive}, making this approach a main research direction for deep learning probabilistic models \cite{lee2022national}. 

Hategan et al. proposed a weighted ensemble model that dynamically integrates machine learning models and sky image–based models for real-time photovoltaic (PV) power forecasting; however, due to insufficient exploitation of inter-data relationships, the prediction accuracy remains low \cite{hategan2025short}. To enhance ultra-short-term PV power forecasting performance, Ma et al. developed an ensemble model based on six individual models using a stacking strategy; nevertheless, the complexity of the model results in substantial computational resource consumption \cite{ma2025research}. Zhou et al. proposed the AM-TCN-BiLSTM model, which combines an attention mechanism, temporal convolutional network (TCN), and bidirectional long short-term memory (BiLSTM) network, and further employs the Rime-ice optimization method to improve PV power forecasting accuracy. However, this approach directly utilizes CEEMDAN decomposition results, leading to suboptimal generalization capability \cite{zhou2025combined,11124293}. Al-Dahidi et al. applied Bayesian optimization to the LSTM-CNN model, improving forecasting accuracy but failing to validate its engineering applicability, such as real-time inference latency and hardware deployment requirements \cite{al2025techno}. Zhang et al. learned temporal patterns from both local and global perspectives and integrated features at multiple scales, proposing a multiscale network with mixed features; however, the limited exploitation of underlying data relationships renders the model highly sensitive to meteorological conditions \cite{zhang2025multiscale}. Li et al. proposed a PV power forecasting approach based on multiscale integration of deep learning models, incorporating wavelet convolution, the Reformer architecture, and Kolmogorov–Arnold Networks (KAN) to capture finer cross-features. Nonetheless, without feature fusion, the directly utilized features lead to limited robustness \cite{li2025novel}. Wang et al. introduced a method combining a long–short-term cross-attention mechanism with physics-informed neural networks to extract temperature features for forecasting, but its performance across different hardware platforms was not assessed \cite{wang2025accurate}. These multi-model ensemble approaches forecast future PV power output by extracting and integrating PV-related features; however, they fail to fully explore the intrinsic relationships among data. For instance, they overlook how meteorological variables physically influence PV generation, which data components affect short-term output, and which influence long-term output.

On the other hand, extensive research has demonstrated that using signal decomposition techniques can handle non-stationarity in data and improve prediction accuracy \cite{husein2024towards}. Data decomposition techniques that have been widely used to date include Empirical Mode Decomposition (EMD) \cite{wang2023accurate,wu2022deterministic,zhang2025short}, CEEMDAN \cite{liang2025short,tang2025short,zhou2025combined,camacho2025short}, Wavelet Transform (WT) \cite{li2021photovoltaic,zhang2023short,arseven2025novel}, Variational Mode Decomposition (VMD) \cite{zhang2022novel,zhai2025photovoltaic,liu2025photovoltaic}, and other techniques. However, WT may require computationally expensive parameter tuning, VMD requires pre-determining the number of modes, and EMD suffers from mode mixing and error accumulation during the decomposition process. These issues hinder the improvement of generalization performance of signal decomposition techniques and impede practical deployment. In contrast, CEEMDAN, as an enhanced version of Ensemble Empirical Mode Decomposition, greatly improves practical application prospects by adding adaptive white noise to the signal at each decomposition stage. It achieves more precise signal decomposition of the original signal without significantly increasing computational load, with minimal reconstruction error. Research has shown that in the field of time series prediction, CEEMDAN can significantly improve the accuracy and robustness of model prediction results \cite{jiang2021forecasting}.

To date, existing literature has not considered the generalization performance issues of CEEMDAN signal decomposition technique across different datasets, nor has it combined the results of signal decomposition with actual weather features for processing. Considering these two aspects would provide significant assistance in improving the accuracy of PV prediction and the generalization performance of the model.

In this study, to enhance the model's generalization performance, we reconstruct the IMFs obtained from CEEMDAN decomposition, reconstructing them into one high-frequency component and one low-frequency component based on frequency thresholds, and use the reconstructed high and low frequency components for PV power generation prediction. During the model training phase, to improve prediction accuracy, width penalty terms are added to the loss function to penalize parts that exceed thresholds.
The main contributions of this paper are as follows:

\begin{itemize}
    \item[1.]  Depending on the frequency, the PV power generation data after CEEDMA decomposition are reconstructed into a high-frequency component and a low-frequency component, improving the generalization performance and prediction capability across different datasets.

    \item[2.] Add width constraints to the loss function of the EQN network, penalizing interval widths that exceed preset thresholds to ensure reasonableness of prediction intervals.

    \item[3.] Designed reasonable feature extraction and learning models to predict ultra-short-term PV power generation, compared to five benchmark models in four datasets, and validated the effectiveness, practicality and generalization performance of the method.
\end{itemize}

\section{Prorposed method}
\begin{figure}[!htb]
\centering
\includegraphics[width=1.0\columnwidth]{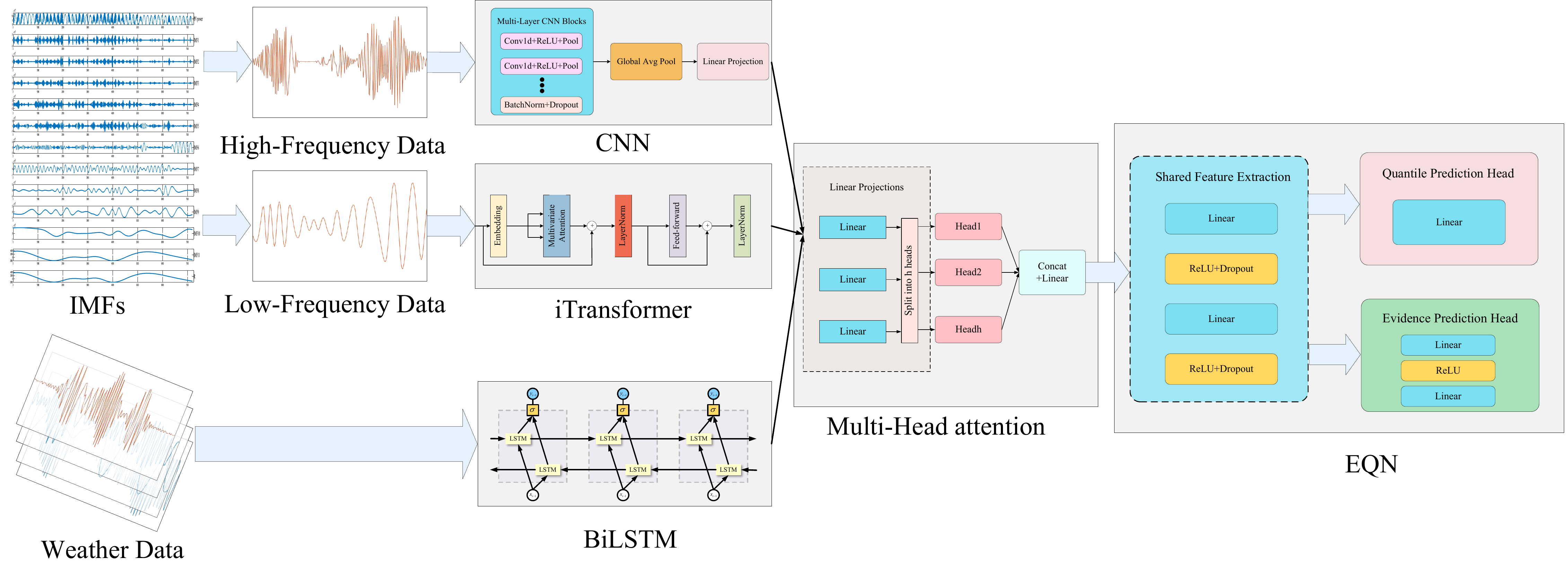}
\caption{Flowchart of CEEDMA-Multi\_nets-EQN. }\label{Fig:mymethod}
\end{figure}

The forecasting method proposed in this paper is illustrated in Fig. \ref{Fig:mymethod}. The PV power generation data are decomposed using CEEMDAN and subsequently reconstructed into a high-frequency component and a low-frequency component. The high-frequency PV components are extracted via CNN, the low-frequency components via iTransformer, and the meteorological features via BiLSTM. Subsequently, the features from these three components are fused using a multi-head attention mechanism, followed by probabilistic forecasting with the EQN network. The specific details are as follows.

\subsection{Missing value handling}\label{Missing value handling}
In the original dataset, there are issues with missing data, which can be categorized into two types: the first is the complete absence of data, and the second is data filled with N/A values to indicate missing entries. Data loss can be classified into the following three situations
\begin{itemize}
    \item[1.] When a small amount of data is missing at the beginning or end of the daily data record, and the contextual data in the vicinity are all zero with a gradual change when PV data recording begins, we fill in the missing data with zero.
    \item[2.] When the context surrounding the missing data contains normal data records and the amount of missing data is only one or two data points, we use linear interpolation to fill in the missing portion. Assuming the contextual data points are \(x_1\) and \(x_2\), the missing data can be calculated as follows:
    \begin{equation}
        x_{i} = x_1 + i\times\frac{(x_2 - x_1)}{n}
    \end{equation}
    where $x_{i}$ represents the $i$th missing data, $n$ is the number of missing data.
    \item[3.] If the context surrounding the missing data consists of normal records but the missing data segment is too long, i.e., the number of missing data points is three or more, we consider the confidence in data supplemented by linear interpolation to be too low. The data of this day will not be used
\end{itemize}

\subsection{CEEMDAN}\label{CEEMDAN}
CEEMDAN multiplies a white noise sequence by a scaling factor based on the characteristics of the original signal to generate adaptive white noise. By incorporating this adaptive white noise, the complex signal is decomposed into a series of intrinsic mode functions (IMFs) and a residual component, serving as an improved version of EMD. CEEMDAN employs adaptive noise at each stage, effectively reducing the mixing of different frequency components and yielding IMFs and residual signals with near-zero reconstruction error. The general steps are as follows:

1. For the PV power sequence \(x(t)\), generate adaptive white noise
\begin{equation}
    \boldsymbol{n}_i(t) = \epsilon \delta_x \omega_i(t)
\end{equation}
where $\delta_x$ is the standard deviation of $x(t)$, $\omega_i(t)$ is the $i$-th standard white noise, and $\epsilon$ is the noise intensity factor.

2. Add noise to the signal/residual:
\begin{equation}
\boldsymbol{x}_k^{(i)}(t) = R_{k-1}^{(i)}(t) + \boldsymbol{n}_i(t)
\end{equation}
Decompose $\boldsymbol{x}_k^{(i)}(t)$ via EMD to obtain the $k$-th IMF, and update the residual:
\begin{equation}
\text{IMF}_k^{(i)}(t) = \text{EMD}\!\left(\boldsymbol{x}_k^{(i)}(t)\right),\quad  
R_k^{(i)}(t) = \boldsymbol{x}_k^{(i)}(t) - \text{IMF}_k^{(i)}(t)
\end{equation}

Repeat until the residual is no longer decomposable.

3. Average $N$ realizations to obtain the final IMFs and residual:
\begin{equation}
\begin{array}{c}
     \text{IMF}_k(t) = \frac{1}{N} \sum_{i = 1}^{N} \text{IMF}_k^{(i)}(t)\\
     R(t) = \frac{1}{N} \sum_{i = 1}^{N} R_K^{(i)}(t)
\end{array}
\end{equation}
This yields $K$ IMFs and one residual component.

\subsection{IMFs data reconstruction}\label{imf}
After decomposing the PV power data using CEEMDAN, \(K\) IMFs and one residual component are obtained. However, the results of CEEMDAN decomposition depend on factors such as the complexity and length of the input data. For instance, data with complex frequency components typically yield more IMFs, whereas relatively stationary data yield fewer IMFs. Consequently, the number of IMFs generated from different datasets may vary significantly, which can severely impair generalization capability. Therefore, further processing of the CEEMDAN decomposition results is necessary to enhance generalization performance.

In real scenarios, PV power generation is affected by various conditions that occur at different frequencies. For example, cloud cover occurs much more frequently than diurnal cycles, and the appearance of instantaneous shadows is much more frequent than changes in solar elevation angle. Therefore, we categorize the decomposed data into low-frequency and high-frequency groups based on frequency. The high frequency group corresponds mainly to rapid cloud movement, atmospheric refraction changes, and dew evaporation, while the low frequency group corresponds primarily to diurnal cycles and seasonal variations. 

Each IMF after the CEEMDAN decomposition can be expressed as follows:
\begin{equation}
    IMF_i={x(n), n=0,1,2,\cdots,N-1}, i=1,2,\cdots,K
\end{equation}
where N is the length of IMF, $f_s$ is the sampling frequency.

For each $IMF_i,i=1,2,\cdots,K$, applying Fast Fourier transform (FFT) gives
\begin{equation}
    X(k) = \sum_{n=0}^{N-1} x(n) e^{-j\frac{2\pi kn}{N}}, \quad k = 0, 1, 2, ..., N-1
\end{equation}

Its power spectral density is
\begin{equation}
    S(k) = |X(k)|^2 = X(k) \cdot X^*(k)
\end{equation}
where $X^*(k)$ is  the complex conjugate of $X(k)$.

Due to the conjugate symmetry of the FFT of real signals, the dominant frequency can be defined as the frequency corresponding to the maximum value of the power spectrum in the positive frequency portion
\begin{equation}
    \begin{array}{c}
        f_{pos}(m) = f(m), \quad m = 1, 2, ..., \lfloor N/2 \rfloor \\
        S_{pos}(m) = S(m), \quad m = 1, 2, ..., \lfloor N/2 \rfloor \\
        f_{dom,i}= f_{pos}(m_{max})
    \end{array}
\end{equation}
where \(m_{max} = \arg\max_{m} S_{pos}(m)\), \(f_{dom,i}\) id the dominant frequency of $i-$th IMF. 

However, relying solely on the dominant frequency for partitioning IMFs is susceptible to noise interference, particularly when the dominant frequency is high. To reduce the impact of noise on IMF reconstruction, we calculate the frequency centroid to assist in the reconstruction process. The frequency centroid physically represents the center point of the spectral energy distribution, and, in the time domain, it reflects the average oscillation rate of the signal. A higher frequency centroid indicates that the signal contains more high-frequency components, whereas a lower frequency centroid suggests that it contains more low-frequency components. The frequency centroid is defined as follows:
\begin{equation}
    f_{cen,i} = \frac{\sum_{m=1}^{\lfloor N/2 \rfloor} f_{pos}(m) \cdot S_{pos}(m)}{\sum_{m=1}^{\lfloor N/2 \rfloor} S_{pos}(m)}
\end{equation}
where $f_{cen,i}$ is the frequency centroid of $i-$th IMF.

When grouping IMFs, if the frequency centroid differs from the dominant frequency by more than 10\%, we consider the current dominant frequency to be the frequency centroid, i.e.
\begin{equation}
    f_{dominant}=\left\{ \begin{array}{cc}  
             f_{cen,i}, &  f_{cen,i}<0.9 f_{dom,i}\\  
             f_{dom,i}, & 0.9 f_{dom,i}\leq f_{cen,i}\leq 1.1f_{dom,i}\\  
             f_{cen,i}, &   1.1f_{dom,i}< f_{cen,i}
             \end{array}  
\right.  
\end{equation}

Set the threshold frequency \(f_{high}\), and group all IMFs according to the threshold frequency
\begin{equation}
  Group_i = \begin{cases} 
\text{High-freq} & \text{if } f_{dom,i} > f_{high} \\ 
\text{Low-freq} & \text{if } f_{dom,i} < f_{high} 
\end{cases}  
\label{equ:group}
\end{equation}

After grouping all IMFs, we sum the IMFs in the high-frequency group to form the high-frequency component, and apply the same processing to the IMFs in the low-frequency group. This way, K IMF components can be integrated into one high-frequency component and one low-frequency component, regardless of the value of K.

\begin{equation}
    \begin{array}{c}
         S_{high}(t) = \sum_{i \in I_{high}} IMF_i(t)  \\
          S_{low}(t) = \sum_{i \in I_{low}} IMF_i(t) +R(t)
    \end{array}
    \label{equ:group_result}
\end{equation}
where \(I_{high}\) and \(I_{low}\) are the index sets of IMFs corresponding to the high-frequency and low-frequency groups, respectively, and \(R(t)\) is the residual term.

After the IMFs data integration, regardless of the number of IMFs resulting from the CEEMDAN decomposition, only the high-frequency and low-frequency signals need to be trained during model training, which will significantly improve the method's generalization performance.

\subsection{CNN}\label{CNN}
CNN is a deep learning model specifically designed for processing sequential data. Its core concept involves automatically extracting local features and global patterns from time series or one-dimensional signals through convolutional operations, effectively capturing spatio-temporal dependencies. It is particularly well-suited for handling sequential data with local correlations. The CNN feature extractor consists of several core components, including convolutional layers, pooling layers, batch normalization layers, and fully connected layers.

\begin{itemize}
    \item [1.] Convolutional Layer, Captures local features via
    \begin{equation}
        y(i) = \sigma\left(\sum (w(k) \times x(i+k)) + b\right)
    \end{equation}
   where $w$ is the convolution kernel, $b$ the bias, and $\sigma$ the activation function. ReLU is applied to introduce nonlinearity, enhance feature extraction, and mitigate overfitting.
   
   \item[2.] Pooling Layer, Downsamples convolutional outputs to remove redundancy, reduce dimensionality, and retain key information, improving efficiency and trend recognition.
   
   \item[3.] Batch Normalization and Dropout, Batch Normalization accelerates convergence and mitigates internal covariate shift, while Dropout (10\%) prevents overfitting, boosting generalization.
   
   \item[4.]  Global Feature Aggregation, Adaptive Average Pooling aggregates global features, followed by a linear projection layer to standardize output dimensions for subsequent modules.
\end{itemize}
\subsection{iTransformer}\label{iTransformer}
It has been demonstrated in various tasks that models based on the Transformer architecture excel at handling long-term dependencies in time series data. The self-attention mechanism primarily focuses on modeling dependencies in the temporal dimension while overlooking the complex cross-correlations and dynamic relationships among multiple variables. Additionally, during layer normalization and feature processing, the gradual merging of multivariate representations introduces interaction noise, further hindering the capture of critical variable correlations. The iTransformer independently embeds the complete time series of each variable as a variable token, rather than fusing multivariate time tokens. This allows the attention mechanism to directly model correlations between variables.The process of feature extraction using iTransformer is as follows:
\begin{itemize}
    \item [1.] Dimension Rearrangement 
    For the input \(\mathbf{X} \in \mathbb{R}^{B \times T \times N}\), perform a dimension transformation: \(\mathbf{X}' = \text{rearrange}(\mathbf{X}) \in \mathbb{R}^{B \times N \times T}\), where the function \(\text{rearrange}\) is used to reorder the dimensions of the tensor.
    \item[2.] Variable Token Embedding
    \begin{equation}
        \mathbf{Z}^{(0)} = \text{LayerNorm}(\text{Linear}(\mathbf{X}')) \in \mathbb{R}^{B \times N \times d}
    \end{equation}
   where \(\text{Linear}(\cdot)\) is a fully connected layer function, and \(\text{LayerNorm}(\cdot)\) is a normalization layer function.
   \item[3.] Transformer Layer.  For each layer \(l = 1, 2, \ldots, L\): 
   \begin{equation}
       \mathbf{Z}^{(l)} = \text{LayerNorm}(\mathbf{Z}^{(l-1)} + \text{FFN}(\text{LayerNorm}(\mathbf{Z}^{(l-1)} + \text{Attention}(\mathbf{Z}^{(l-1)}))))
   \end{equation}
   where \(\text{Attention}(\cdot)\) is the self-attention mechanism used to compute correlations between variables, and \(\text{FFN}(\cdot)\) is the feedforward neural network.
   \item[4.]  Feature Extraction Output 
   \begin{equation}
       \mathbf{Features} = \mathbf{Z}^{(L)} \in \mathbb{R}^{B \times N \times d}
   \end{equation}
\end{itemize}
\subsection{BiLSTM}\label{BiLSTM}
LSTM is designed to process and predict significant events in time series data with relatively long intervals and delays. The core of LSTM lies in three gating mechanisms: the input gate, forget gate, and output gate, which employ the sigmoid function to regulate the extent of information flow.
\begin{equation}
    \begin{array}{c}
         f_t = \sigma(W_f \cdot [h_{t-1}, x_t] + b_f)\\
         i_t = \sigma(W_i \cdot [h_{t-1}, x_t] + b_i)\\
         \tilde{C}_t = \tanh(W_u \cdot [h_{t-1}, x_t] + b_u)\\
         O_t = \sigma(W_o \cdot [h_{t-1}, x_t] + b_o)\\
         C_t = f_t \odot C_{t-1} + i_t \odot \tilde{C}_t\\
         h_t = O_t \odot \tanh(C_t)\\
    \end{array}
\end{equation}
where $f_t, i_t, O_t$ are gate activations, $\tilde{C}_t$ is the candidate cell state, $C_t$ the cell state, $h_t$ the hidden state, and $\odot$ denotes element-wise multiplication.

\begin{figure}[!htb]
\centering
\includegraphics[width=1.0\columnwidth]{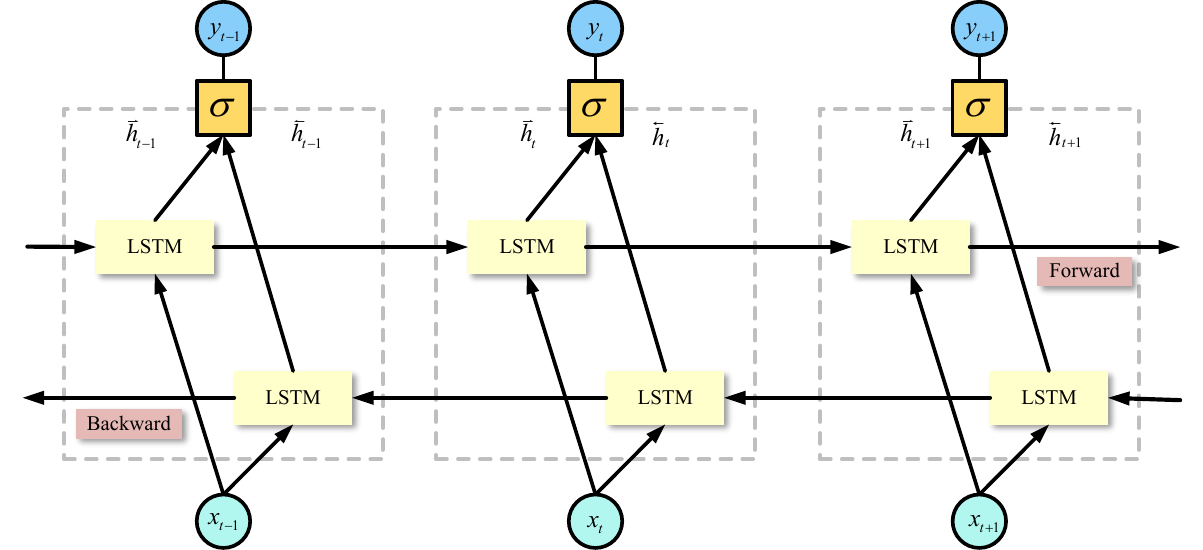}
\caption{The structure of BiLSTM. }\label{Fig:bilstm}
\end{figure}

BiLSTM addresses the limitation that traditional LSTM cannot feed back information from later data to earlier data for judgment by introducing a backward-propagating LSTM layer in addition to the traditional LSTM. At each time step, the BiLSTM output is a concatenation of the forward and backward LSTM outputs, as shown in Fig. \ref{Fig:bilstm}.
By stacking multiple BiLSTM layers, the model can learn feature representations at different levels, achieving progressive feature extraction from simple to complex.



\subsection{Multi-head Attention}\label{Attention}
The attention mechanism mimics the human brain's information processing by assigning higher weights to important information and lower weights to less significant information, enhancing the model's focus on critical data. The mathematical formulation of the multi-head attention mechanism is as follows:

\begin{equation}
\begin{array}{c}
    \text{Attention}(Q, K, V) = \text{softmax}\left(\frac{QK^T}{\sqrt{d_k}}\right)V \\
    \text{head}_i = \text{Attention}(QW_i^Q, KW_i^K, VW_i^V) \quad i \in [1, N] \\
    \text{MultiHead}(Q, K, V) = \text{Concat}(\text{head}_1, \ldots, \text{head}_h)W^O
\end{array}
\end{equation}
where \(\text{MultiHead}(\cdot)\) refers to the multi-head attention mechanism, \(Q\), \(K\), and \(V\) represent the query, key, and value vectors, respectively, \(\text{head}_i\) is the \(i\)-th prediction head among \(N\) heads, and \(W_i^Q\), \(W_i^K\), \(W_i^V\), and \(W^O\) are learnable weight matrices.
\subsection{EQN}\label{EQN}
\begin{figure}[!htb]
\centering
\includegraphics[width=1.0\columnwidth]{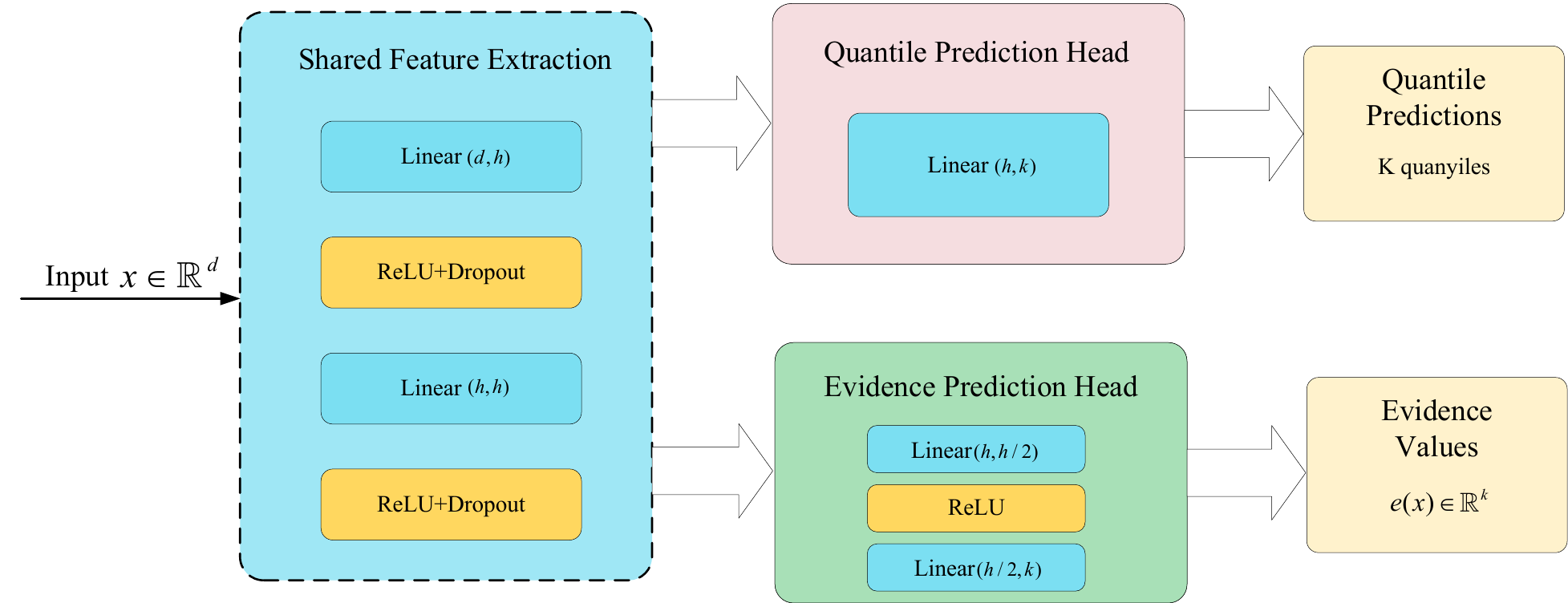}
\caption{The structure of EQN. }\label{Fig:EQN}
\end{figure}

The EQN proposed by Sensoy et al. (2018) \cite{sensoy2018evidential}, combines quantile regression with evidential learning to jointly address prediction accuracy and uncertainty quantification in regression tasks. Its architecture consists of a shared feature extractor, a quantile prediction head, and an evidence prediction head, as shown in Fig \ref{Fig:EQN}

The primary purpose of the shared feature extraction layer is to extract high-level abstract feature representations from raw input data, which serve both the quantile prediction and evidence estimation tasks. By leveraging shared feature representations, the network capitalizes on the intrinsic correlations between these two related tasks, enhancing overall learning efficiency and generalization ability. 

Quantile prediction learns conditional distributions at predefined quantiles:
\begin{equation}
    \hat{q}_\tau(x) = f_\theta^{(q)}(h(x))
\end{equation}
where $f_\theta^{(q)}$ outputs the $\tau$-th quantile.

The evidence head estimates prediction confidence via evidence strength:
\begin{equation}
    e_{\tau,i} = f_\theta^{(e)}(h(x_i)), \quad \alpha_{\tau,i} = e_{\tau,i} + 1
\end{equation}

EQN quantifies epistemic uncertainty as the inverse of evidence:
\begin{equation}
    u_{\text{epistemic}} = \frac{1}{e_\tau + \epsilon}
\end{equation}
and aleatoric uncertainty as the quantile range:
\begin{equation}
    u_{\text{aleatoric}} = \hat{q}_{\tau_{idx_{\text{upper}}}}(x) - \hat{q}_{\tau_{idx_{\text{lower}}}}(x)
\end{equation}
where $\alpha$ is the tail proportion.

By producing both pointwise quantile estimates and reliable uncertainty measures in a single forward pass, EQN improves robustness under distribution shifts and data scarcity.

To make the predicted values approach the true quantiles while preventing the model from being overconfident, based on quantile regression theory and Evidential Learning principles, the quantile loss is defined as:
\begin{equation}
    \begin{array}{c}
         \mathcal{L}_{\text{quantile}} = \frac{1}{N} \sum_{i=1}^{N} \sum_{j=1}^{Q} \rho_{\tau_j}(y_i - \hat{q}_{i,j})\\
         \rho_{\tau}(u) = u \cdot (\tau - \mathbb{I}_{u < 0}) = \begin{cases}
\tau \cdot u, & \text{if } u \geq 0 \\
(\tau - 1) \cdot u, & \text{if } u < 0
\end{cases} 
    \end{array}
\end{equation}
where $N$ is the batch size, $Q$ is the number of quantiles, $y_i$ is the true value of the $i$-th sample, $\hat{q}_{i,j}$ is the $j$-th quantile prediction of the $i$-th sample, $\tau_j$ is the $j$-th quantile level, and $\mathbb{I}_{u < 0}$ is the indicator function.

The regularization loss is defined as follows:
\begin{equation}
  \mathcal{L}_{\text{evidence}} = \frac{1}{N} \sum_{i=1}^{N} \sum_{j=1}^{Q} e_{i,j}  
\end{equation}
where $e_{i,j}$ is the evidence value of the $j$-th quantile for the $i$-th sample.

The loss function can be written as
\begin{equation}
    \mathcal{L}_{\text{EQN}} = \lambda_1\mathcal{L}_{\text{quantile}} + \lambda_2 \mathcal{L}_{\text{evidence}}
    \label{ori_loss}
\end{equation}
where $\lambda_1$ is the quantile regularization coefficient, $\lambda_2$ is the evidence regularization coefficient.

However, the loss function \eqref{ori_loss} encourages accurate prediction of each quantile but does not directly constrain interval width. During the training process, the model will improve coverage by expanding the prediction interval, making $\mathcal{L}_{\text{quantile}}$ reach a low value, and the loss function will decrease accordingly. As training progresses, if the prediction interval width is not constrained, the prediction interval will become increasingly wide, and the loss function will fall into local optima, ultimately leading to overly wide prediction intervals, reducing the value of prediction results, and even making the decision support value of predictions almost zero. Therefore, it is necessary to correct this tendency through width constraints, forcing the model to pursue more compact and valuable prediction intervals while ensuring coverage.

To limit interval width, maximum widths are set for all thresholds. In the initial stages of training, the model tends to generate wide intervals to ensure coverage, which leads to high width loss, and the loss function penalizes parts that exceed thresholds to constrain interval width. As training progresses, the model learns to narrow intervals while maintaining coverage, ultimately improving the accuracy of quantile predictions and achieving an optimal balance between accuracy and width. This width constraint forces the model to learn more precise and valuable uncertainty quantification methods while ensuring probability calibration.
\begin{equation}
    \mathcal{L}_{\text{width}} = \frac{1}{W} \sum_{k} w_k \cdot \frac{1}{N} \sum_{i=1}^{N} \max(\Delta q_{i,k} - \theta_k)
\end{equation}
where $w_k$ is the weight of the $k$-th confidence interval, $\theta_k$ is the maximum width threshold of the $k$-th confidence interval, $W = \sum_k w_k$ is the normalization constant.

In summary, the loss function designed in this paper is:
\begin{equation}
    \mathcal{L}_{\text{EQN}} = \lambda_1\mathcal{L}_{\text{quantile}} + \lambda_2 \mathcal{L}_{\text{evidence}} + \lambda_3\mathcal{L}_{\text{width}}
\end{equation}
where $\lambda_3$ is the width penalty coefficient.

\subsection{Architecture of the proposed method}

For a PV power generation sequence, this paper decomposes it and reconstructs it into one high-frequency component and one low-frequency component, and predicts future PV power generation through feature extraction. Since different data have different characteristics, this paper combines CEEMDAN, CNN, iTransformer, BiLSTM, and EQN to propose the CEEDMA-Multi\_nets-EQN hybrid prediction model.

The model performs CEEMDAN decomposition on PV power generation data to obtain IMFs containing different frequency components and one residual term. Based on the dominant frequency and frequency centroid of different IMFs relative to the frequency threshold, they are divided into high-frequency IMF groups and low-frequency IMF groups, then reconstructed into one high-frequency component and one low-frequency component. This way, for different datasets, regardless of their data volume, we can always obtain one high-frequency component and one low-frequency component. After standardizing the high-frequency component, low-frequency component, and meteorological data, CNN is used to extract features from the high-frequency component, iTransformer is used to extract features from the low-frequency component, and BiLSTM is used to extract features from the meteorological data. After feature extraction is completed, multi-head attention mechanism is applied to all features for feature fusion, enhancing the model's attention to important features. Finally, the fused features are fed into the EQN network for PV power generation prediction.

\begin{figure}[!htb]
\centering
\includegraphics[width=1.0\columnwidth]{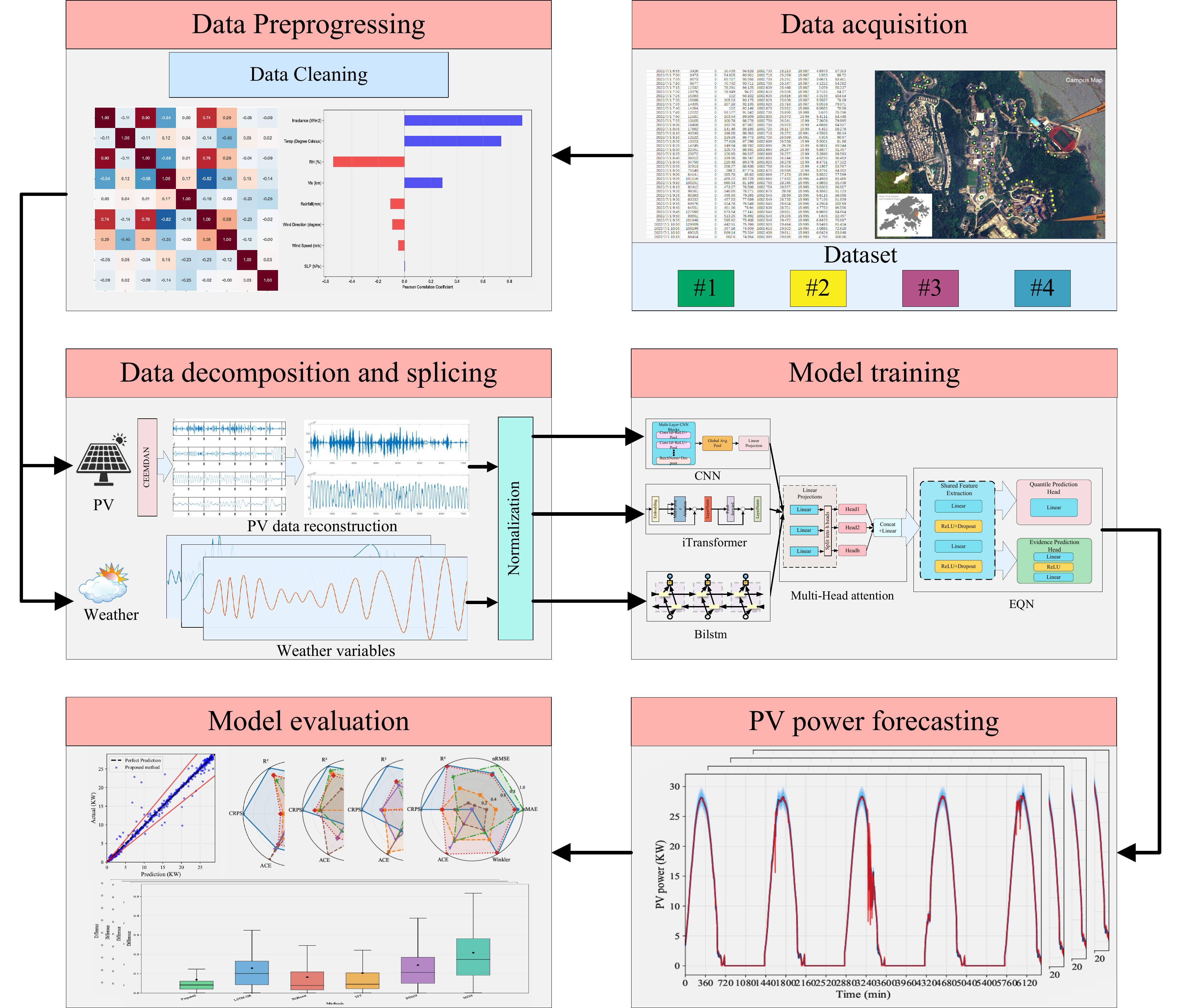}
\caption{Framework of the prediction model proposed in this paper. }\label{Fig:total_program}
\end{figure}

Based on CEEDMA-Multi\_nets-EQN, Fig. \ref{Fig:total_program} depicts the flowchart of PV power generation prediction. Its main process is summarized as follows:
\begin{itemize}
    \item [(a)] Data preprocessing. Use linear interpolation to fill missing values in the middle of the original PV power generation sequence, supplement the missing values at the beginning and end of each day according to conditions, and remove sequences that do not meet the requirements.
    \item [(b)] Data decomposition. By executing CEEMDAN on the completed PV power generation sequence, obtain IMFs containing different frequency components and one residual term.
    \item [(c)] IMF reconstruction. Group according to the different IMFs' dominant frequency and frequency centroid relative to the frequency threshold, and reconstruct into one high-frequency component and one low-frequency component.
    \item [(d)] Sample set construction. Perform standardization processing on the high-frequency component, low-frequency component, and meteorological data, and allocate the processed data into training set, validation set, and test set in a ratio of 0.8:0.1:0.1. Use data from the previous 12 time steps to predict the next time step.
    \item [(e)] Feature extraction. Use CNN, iTransformer, and BiLSTM to extract features from high-frequency components, low-frequency components, and meteorological data, respectively.
    \item [(f)] Feature fusion. Use multi-head attention mechanism to learn the correlations between the three features, allowing features from different sources to mutually attend to and influence each other, discover cross-modal dependencies, and automatically calculate the importance weights of each branch.
    \item [(g)] Model prediction. Based on the fused features, use the EQN network for probabilistic prediction to obtain quantile prediction results and uncertainty estimation.
    \item [(h)] Calculate the loss function. For each sample, its loss includes three parts: quantile loss, evidence regularization loss, and width constraint.
    \item [(i)] Model training. Learn the CEEDMA-Multi\_nets-EQN model by minimizing the loss function on the training set, where the Adam optimizer is used to update the model parameters.
\end{itemize}

\section{Experiment Description and analysis}

\subsection{Dataset description}

\begin{figure}[!htb]
\centering
\includegraphics[width=1.0\columnwidth]{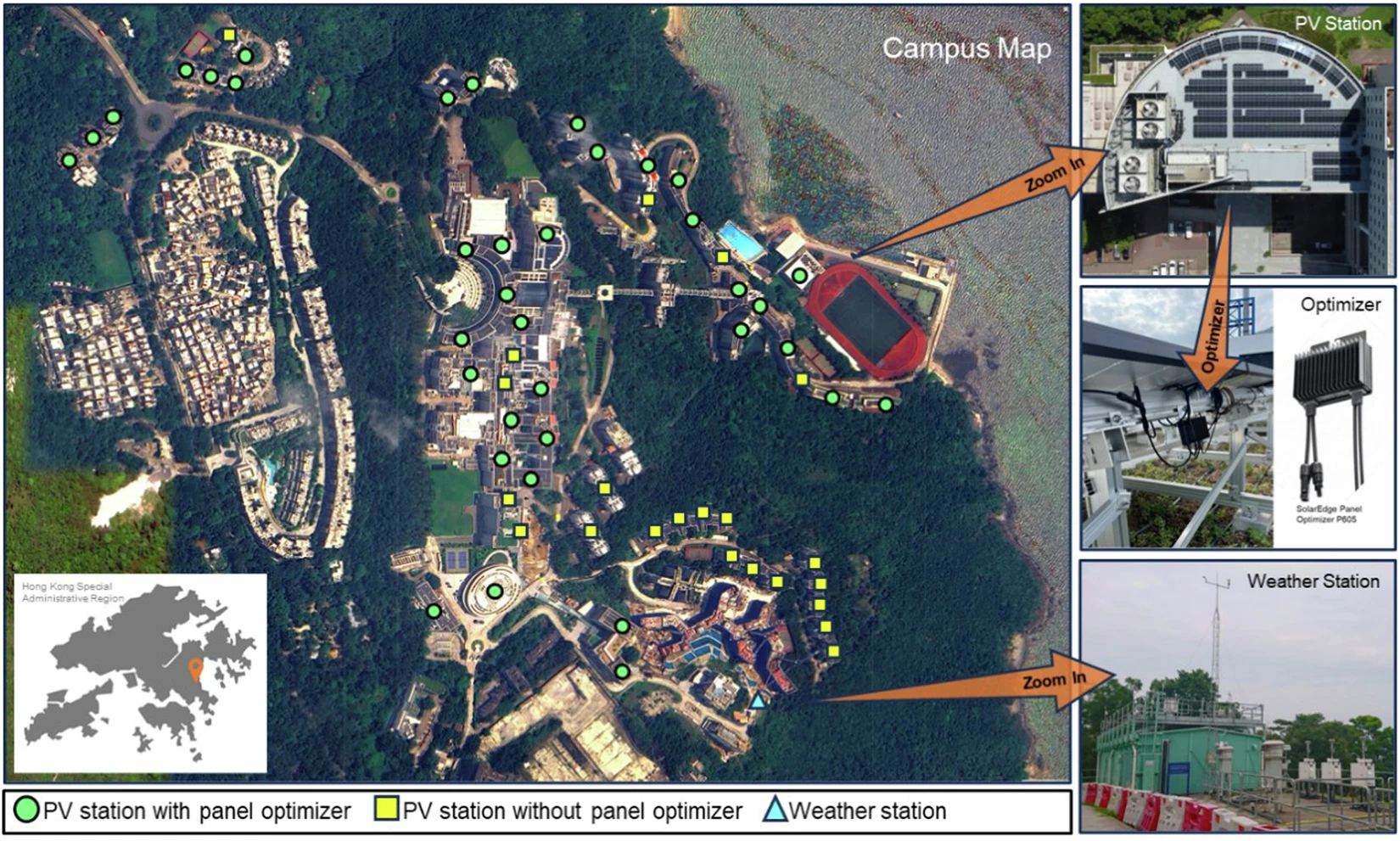}
\caption{Satellite image location map of the PV power station and meteorological station.}\label{Fig:site}
\end{figure}

The data used in this study were obtained from a rooftop solar project managed by the University's Sustainability/Net Zero Office. The project is implemented on the campus of Hong Kong University of Science and Technology, located at 22.3363 ° N latitude and 114.2634 ° E longitude, covering an area of 60 hectares. This project comprises 60 grid-connected rooftop PV power stations distributed in various locations on the campus, incorporating a total of 6,085 PV modules. In addition, it includes a meteorological station for collecting weather data, as illustrated in Fig. \ref{Fig:site}. All PV modules are sourced from JinkoSolar and utilize monocrystalline silicon technology, with power outputs ranging from 365W to 415W. Data are measured every 5 minutes with a measurement device accuracy of ±2.5\%.

Considering the climatic conditions and equipment variations in the Hong Kong region, to fully address the dual challenges in PV power generation forecasting, we selected three months of data from July 1, 2022, to September 30, 2022, from four types of equipment with the highest installation quantities as our original dataset. These four equipment types are: JKM365N-6TL3-V (365W), efficiency: 20.96\% (Dataset \#1); JKM370N-6TL3 (370W), efficiency: 21.0\% (Dataset \#2); JKM390M-6RL3-TV (390W), efficiency: 20.12\% (Dataset \#3); and JKM410N-6RL3 (415W), efficiency: 21.48\% (Dataset \#4). 
\subsection{Correlation analysis}
The original dataset includes eight types of meteorological data that have an impact on PV power generation to varying degrees. To accelerate the model training process and reduce prediction errors, we calculated the Pearson correlation coefficient to conduct a correlation analysis. From this, we selected the meteorological data with the most significant impact on PV power generation as inputs for the model. The Pearson correlation coefficient is a statistical measure of the linear relationship between two continuous variables, with values ranging from -1 to 1, denoted as \(\rho\). Its calculation formula is as follows:
\begin{equation}
\rho  = \frac{\sum_{i=1}^{n} (x_i - \bar{x})(y_i - \bar{y})}{\sqrt{\sum_{i=1}^{n} (x_i - \bar{x})^2 \sum_{i=1}^{n} (y_i - \bar{y})^2}}
\end{equation}
where \(x_i\) and \(y_i\) represent the raw values of meteorological data and PV power generation data, respectively, \(\bar{x}\) and \(\bar{y}\) are the mean values of the meteorological data and PV power generation data, respectively, and \(n\) is the total number of data points. 
\begin{figure}[!htb]
\centering
\includegraphics[width=1.0\columnwidth]{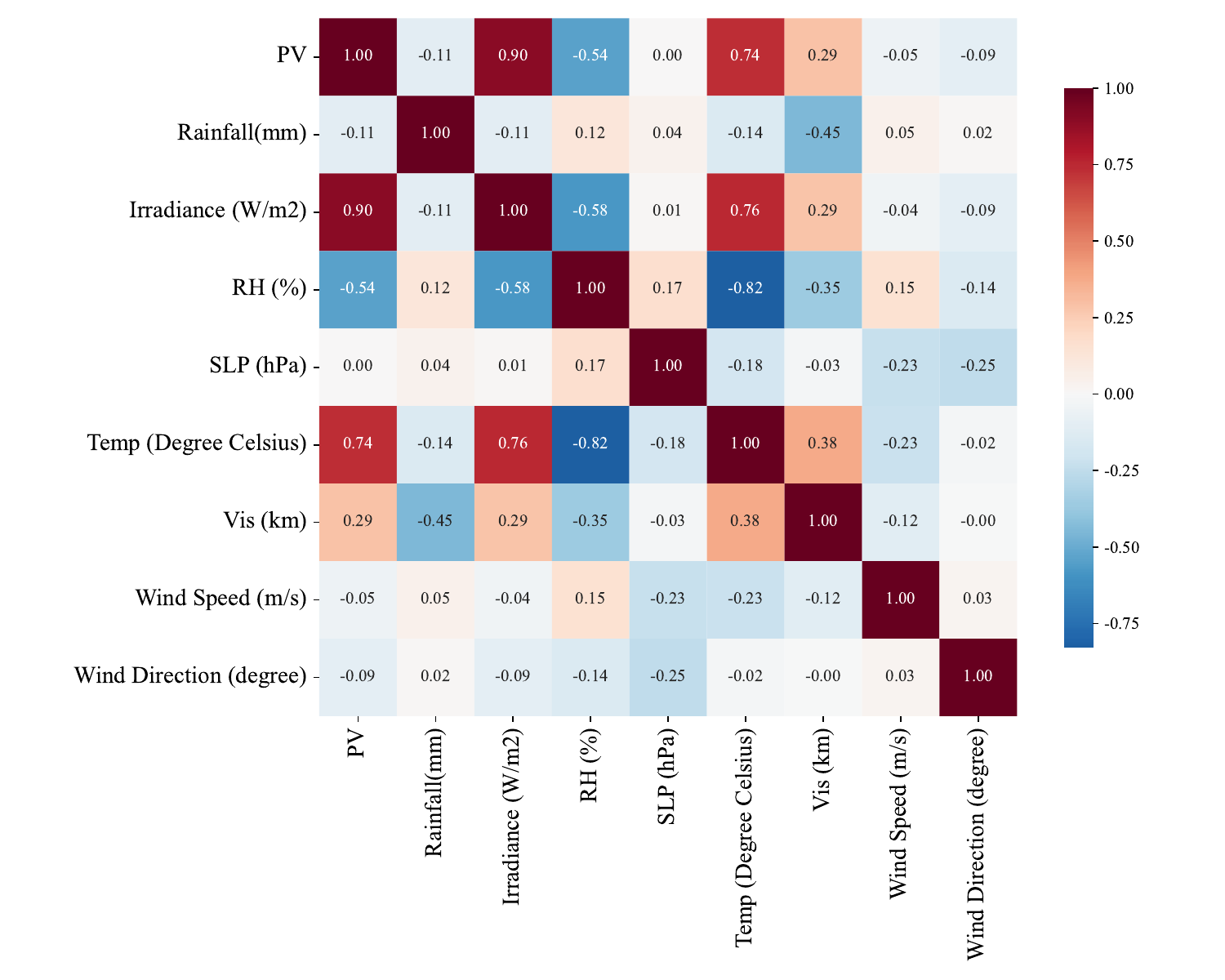}
\caption{Heat map of the variable correlation matrix.}\label{Fig:heatmap}
\end{figure}
\begin{figure}[!htb]
\centering
\includegraphics[width=1.0\columnwidth]{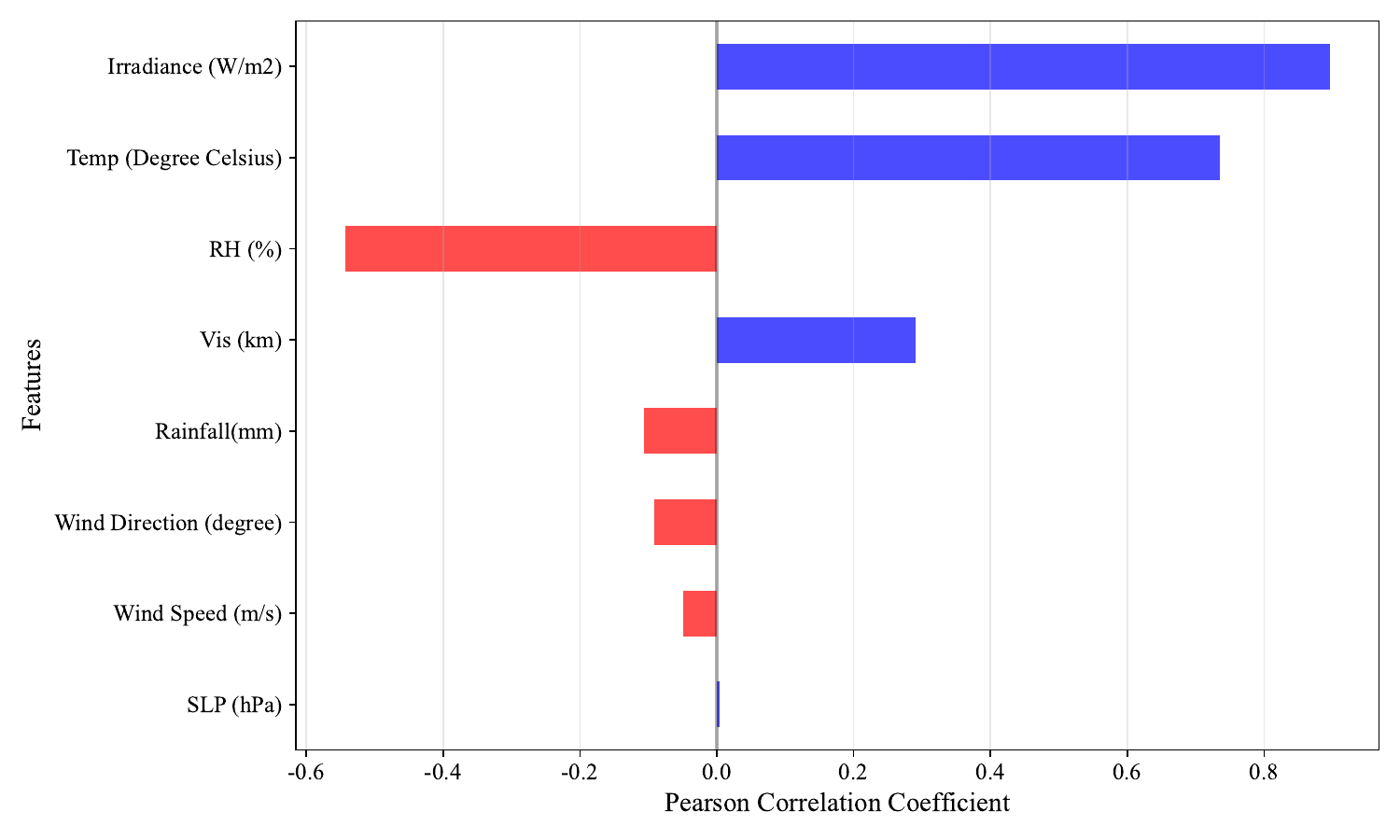}
\caption{Correlation between meteorological data and PV power generation data.}\label{Fig:Correlation Coefficient}
\end{figure}

From Fig. \ref{Fig:heatmap}, it can be observed that irradiance, temperature, humidity, and visibility exhibit relatively large absolute Pearson correlation coefficients with PV power, indicating a more pronounced impact on PV power output. In conjunction with Fig. \ref{Fig:Correlation Coefficient}, these variables can be considered mutually complementary in characterizing weather conditions, while other features show weak or redundant correlations. Therefore, these four variables are selected as the meteorological features.

\subsection{Baselines}
The comparative methods in this study consist of three categories. Quantile Regression Methods include: LSTM-QR, which serves as a baseline for probabilistic forecasting \cite{wang2022approach}, and XGBoost Quantile \cite{chen2016xgboost}, a widely used quantile regression method for probabilistic forecasting in industry. Probabilistic Distribution Modeling Methods include: Mixture Density Network (MDN) \cite{raidoo2022data}, an architecture that combines traditional neural networks with probabilistic density modeling, primarily used to address one-to-many mapping problems and uncertainty modeling; and Beta Distribution NN (BDNN) \cite{fernandez2023short}, specifically designed for modeling Beta distributions within the $[0,1]$ range, which is highly suitable for PV data distributions. Additionally, a specialized time series probabilistic forecasting method, Temporal Fusion Transformer (TFT), proposed by the Google research team in 2020 \cite{lim2021temporal}, is a deep learning architecture designed for multivariate time series forecasting.

\subsection{Model evaluation metrics}
This study evaluates model performance using both deterministic and probabilistic metrics across multiple datasets.

Deterministic metrics include normalized mean absolute error (nMAE), normalized root mean squared error (nRMSE) and the coefficient of determination $R^2$, calculated from the median prediction values. nMAE and nRMSE measure absolute and squared deviations between predictions and observations, normalized to remove scale effects, while $R^2$ indicates the proportion of variance explained by the model:
\begin{equation}
    \mathrm{nRMSE} = \frac{\sqrt{\frac{1}{n} \sum_{t=1}^{n} (y_t - \hat{y}_t)^2}}{\frac{1}{n} \sum_{t=1}^{n} y_t} \times 100\%
\end{equation}
\begin{equation}
    \mathrm{nMAE} = \frac{\frac{1}{n} \sum_{t=1}^{n} |y_t - \hat{y}_t|}{\frac{1}{n} \sum_{t=1}^{n} y_t} \times 100\%
\end{equation}
\begin{equation}
  \mathrm{R}^2 = 1 - \frac{\sum_{t=1}^{n} (y_t - \hat{y}_t)^2}{\sum_{t=1}^{n} (y_t - \bar{y}_t)^2}  
\end{equation}
where \(n\) is the number of data points, \(y_t\) is the actual value of PV power generation, \(\hat{y}_t\) is the predicted PV power generation value, and \(\bar{y}_t\) is the mean of the actual PV power generation values.

Probabilistic metrics include the Continuous Ranked Probability Score (CRPS), the Average Coverage Error (ACE), and the Winkler Score (WS). CRPS evaluates the alignment between predicted probability distributions and observations:
\begin{equation}
  \text{CRPS}(F, y) = \int_{-\infty}^{+\infty} [F(x) - \mathbb{I}(x \geq y)]^2 dx  
\end{equation}
where \(\mathbb{I}(x \geq y)\) is the indicator function, equal to 1 when \(x \geq y\), and 0 otherwise.

ACE measures the deviation between actual and target coverage probabilities:
\begin{equation}
   \text{ACE} = \frac{1}{n} \sum_{i=1}^n \mathbb{I}(y_i \in [L_i, U_i]) - \alpha 
\end{equation}
where \(\alpha\) is the preset confidence level, \([L_i, U_i]\) is the prediction interval for the \(i\)-th sample.

WS combines interval width and under-coverage penalties:
\begin{equation}
    \text{WS} = \frac{1}{n} \sum_{i=1}^n \left[ (U_i - L_i) + \frac{2}{\alpha} \max(L_i - y_i, y_i - U_i, 0) \right]
\end{equation}
where \(\alpha\) is the preset confidence level, \((U_i - L_i)\) represents the interval width, and the second term is a penalty for under-coverage.

Resource usage metrics include Average CPU utilization (ACU), Average Memory Consumption (AMC, MB), Average GPU Memory Utilization (AGMU, MB), Average GPU Utilization Rate (AGUR), Total Training Duration (TTD, min) and Average Epoch Time (AET, s).

\subsection{Parameter tuning}
This study employs the \(Optuna^2\) method to find optimal model parameters within the specified parameter ranges, with the names of the hyperparameters and their corresponding optimal values presented in Tab. \ref{tab:Hyperparameter}. During the training process, the Adam optimizer is used to optimize the parameters, and an early stopping mechanism is introduced to prevent overfitting.
\begin{table*}[htbp]
\centering
\caption{Key hyperparameters of each model.}
\label{tab:Hyperparameter}
\begin{tabular}{>{\centering\arraybackslash}m{3cm}p{10cm}}
\toprule
Model &Hyperparameter\\
\midrule
\multirow{4}{*}{Proposed method} &CNN: filters=64, num-layers=2, kernel-size=3, pooling-layers=2. iTransformer: lookback-len=1, pred-length=1, dim=128, depth=4, heads=8.
EQN: input-size=64, hidden-size=128,num-quantiles=11\\
\cline{2-2}
\multirow{1}{*}{LSTM-QR} &hidden-size=128, num-layers=3, dropout=0.05\\
\cline{2-2}
\multirow{3}{*}{XGBoost} &lookback-window=12,min-child-weight=1, random-state=42, subsample=0.8, colsample-bytree=0.8, max-depth= 8, n-estimators=1500\\
\cline{2-2}
\multirow{2}{*}{TFT} & hidden-size = 64, num-heads = 4, dropout-rate = 0.06, encoder-length = 12\\
\cline{2-2}
\multirow{3}{*}{BDNN} & Transformer-model=128, attention-heads=8, Transformer-layers=2, lstm-hidden=64, dropout=0.1, cnn-filters=64, kernel-size=3\\
\cline{2-2}
\multirow{2}{*}{MDN}& hidden-size=128,   num-components=3,  seq-len=12, dropout=0.2\\
\bottomrule
\end{tabular}
\end{table*}
\section{Experimental results and discussion}
\subsection{Data reconstruction results}
\begin{figure}[!htb]
\centering
\includegraphics[width=1.0\columnwidth]{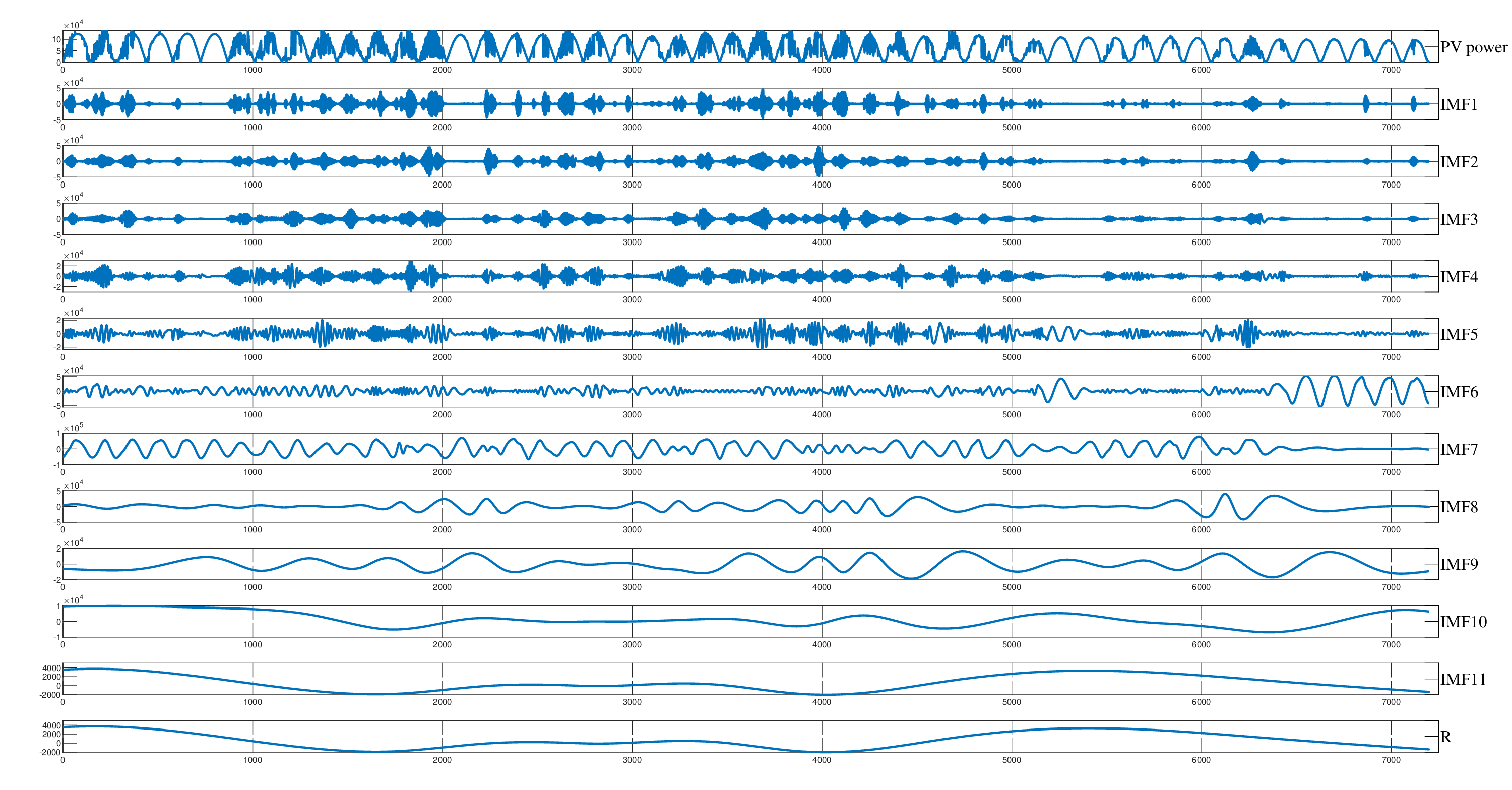}
\caption{CEEMDAN decomposition diagram. }\label{Fig:CEEMDAN}
\end{figure}

In this subsection, we verify whether the data reconstruction method meets the expected requirements. Using Dataset \#1 as an example, the CEEMDAN decomposition yields 11 IMFs and one residual component, as shown in the Fig. \ref{Fig:CEEMDAN}. It is evident from the figure that, as the decomposition progresses, the fluctuations of the sub-series gradually approach stationarity, indicating that the CEEMDAN decomposition effectively reduces the non-stationarity in the PV power generation data.

\begin{figure}[!htb]
\centering
\includegraphics[width=1.0\columnwidth]{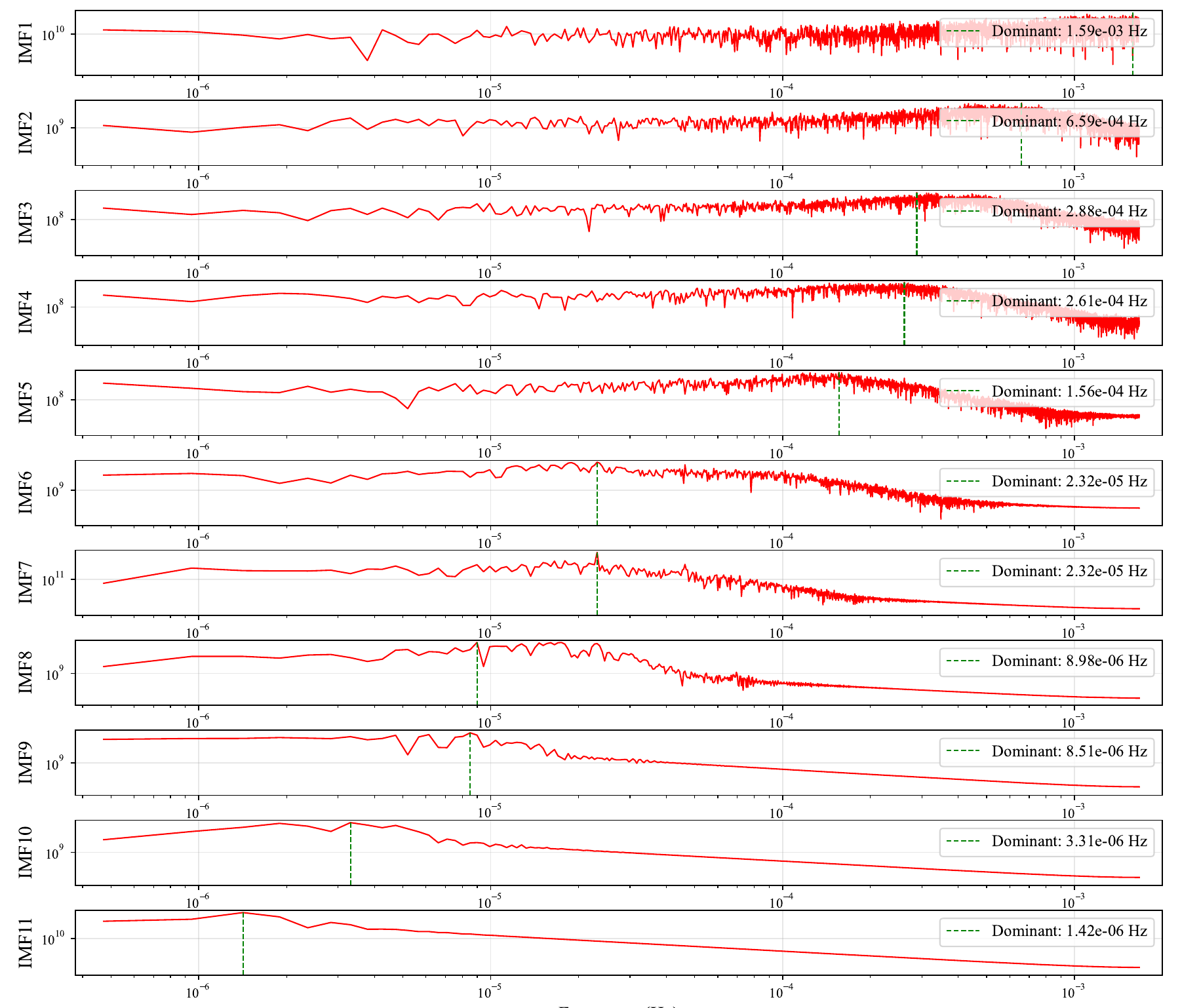}
\caption{FFT results of IMFs for Dataset \#1. }\label{fig:frequency_plot}
\end{figure}

Fig. \ref{fig:frequency_plot} presents the results of the IMF frequency analysis, showing that the CEEMDAN decomposition progressively extracts the frequency components from high to low, layer by layer, from the original PV power generation data. These frequency components correspond to different physical phenomena; for example, the $1.59\times10^{-3}\,\mathrm{Hz}$ component is associated with cloud movement or shadowing from obstacles, while the $2.32\times10^{-5}\,\mathrm{Hz}$ component relates to the semidiurnal patterns of cloud cover.

\begin{figure}[!htb]
\centering
\includegraphics[width=1.0\columnwidth]{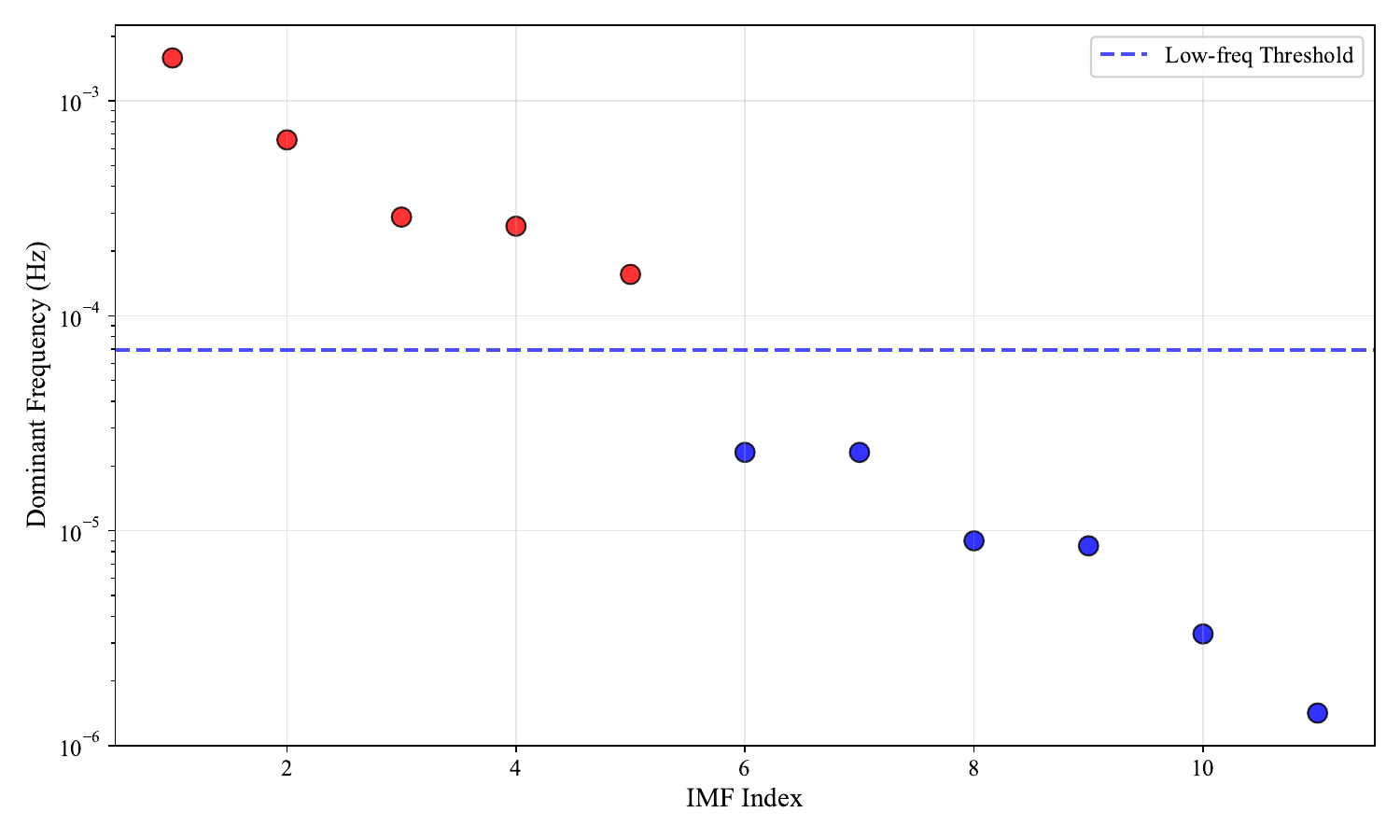}
\caption{The IMF grouping results of IMFs for Dataset \#1. }\label{fig:frequency_Threshold}
\end{figure}

The IMFs obtained from the decomposition are grouped using Equ. \eqref{equ:group}, and the results are shown in the Fig. \ref{fig:frequency_Threshold}.
In the Fig. \ref{fig:frequency_Threshold}, red solid circles represent IMFs classified into the high-frequency group, blue solid circles represent IMFs classified into the low-frequency group, and the blue dashed line represents the grouping frequency threshold. In this paper, we set \(f_{high} = \frac{1}{4\times60 \times 60}\) Hz. It can be clearly seen that for Dataset \#1, the high-frequency group has 5 components, specifically IMF 1-5, and the low-frequency group has 7 components, specifically IMF 6-11 and one residual component.

\begin{figure}[!htb]
\centering
\includegraphics[width=1.0\columnwidth]{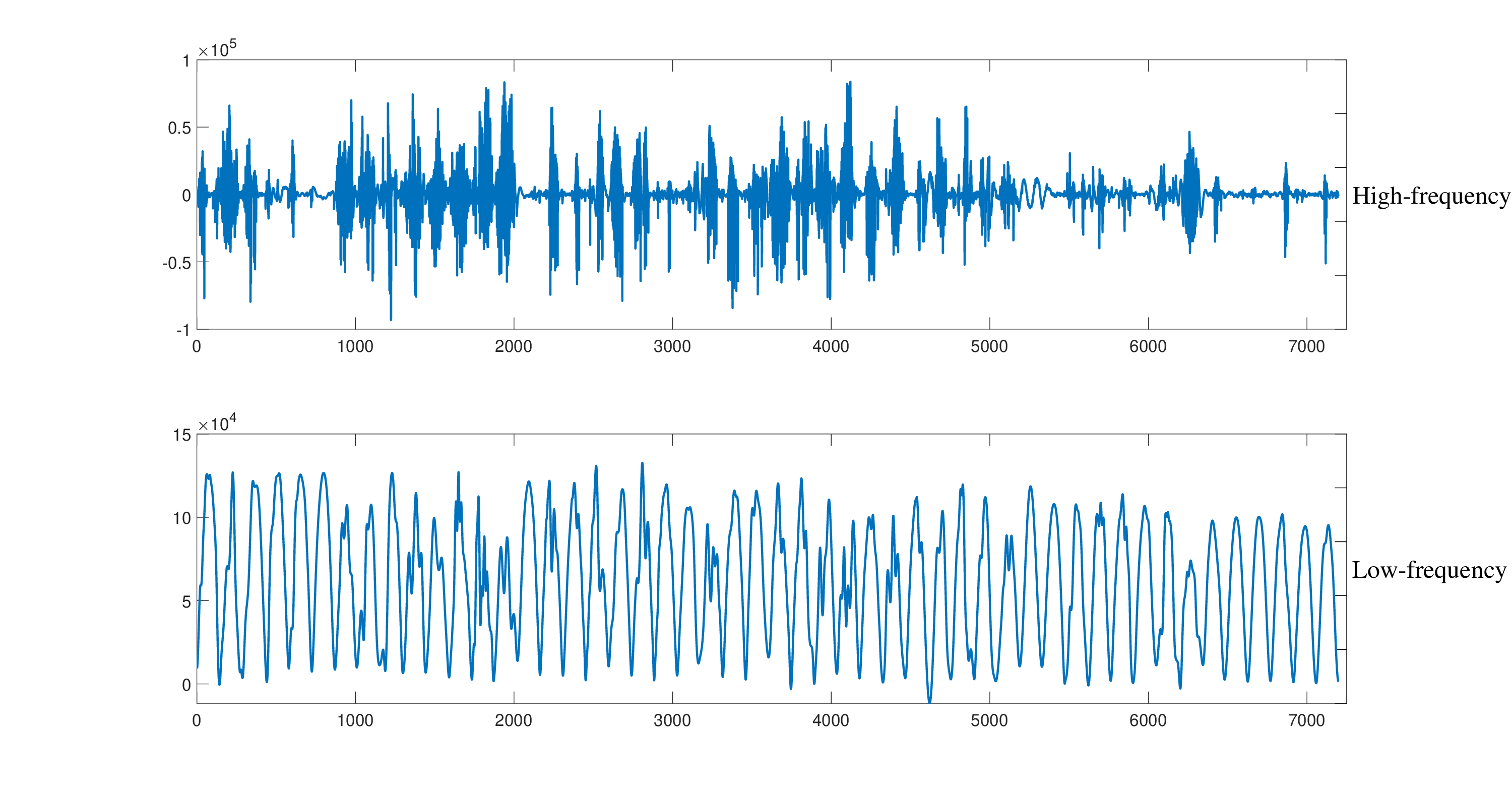}
\caption{IMF reconstruction results. }\label{Fig:frequency group}
\end{figure}

Through Equ. \ref{equ:group_result}, the IMF reconstruction results are shown in Fig. \ref{Fig:frequency group}.
Upon verification, the reconstruction error is $1.38\times e^{-12}$, which is approximately equivalent to measurement noise and can be considered negligible.

\subsection{Comparisons Among Different Baselines}
\begin{figure}[!htb]
\centering
\includegraphics[width=1.0\columnwidth]{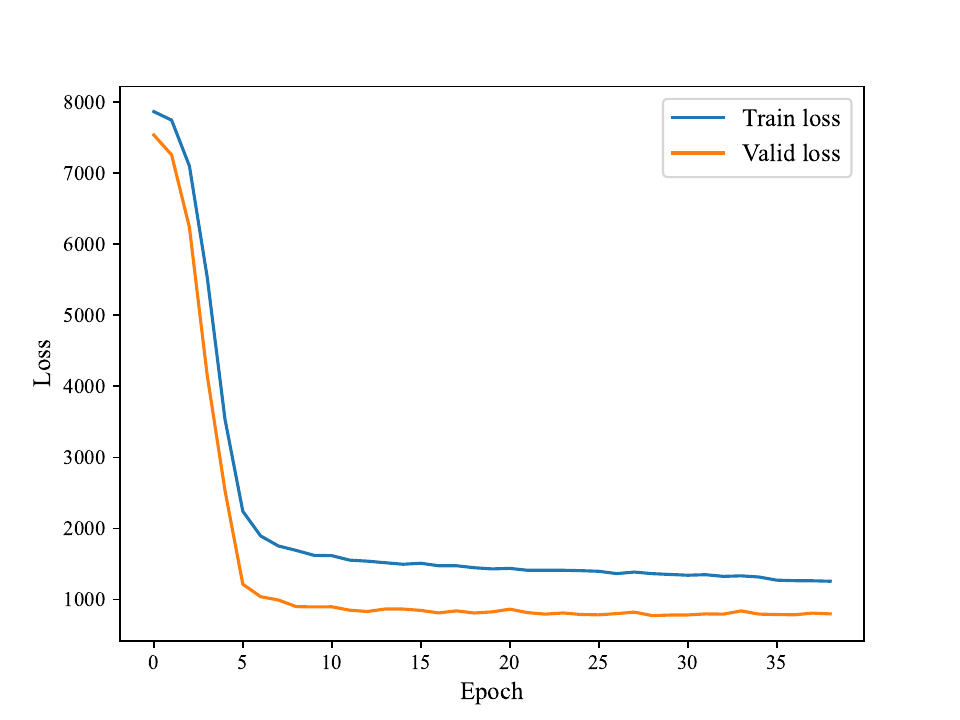}
\caption{Loss curves of CEEDMA-Multi\_nets-EQN  on dataset \#1.}\label{Fig:loss}
\end{figure}

Fig. \ref{Fig:loss} shows the loss curves during the training phase of the proposed method on Dataset \#1. As can be seen from the figure, within the first 5 epochs, both training loss and validation loss drop sharply from approximately 8000 to below 2000, demonstrating that the model has high learning efficiency in the early stage. After 5 epochs, both curves tend to level off. The trends of training loss and validation loss are highly consistent, and the validation error is slightly lower than the training loss, indicating that the model does not exhibit overfitting and has good generalization ability.

\begin{figure}[!htb]
\centering
\includegraphics[width=1.0\columnwidth]{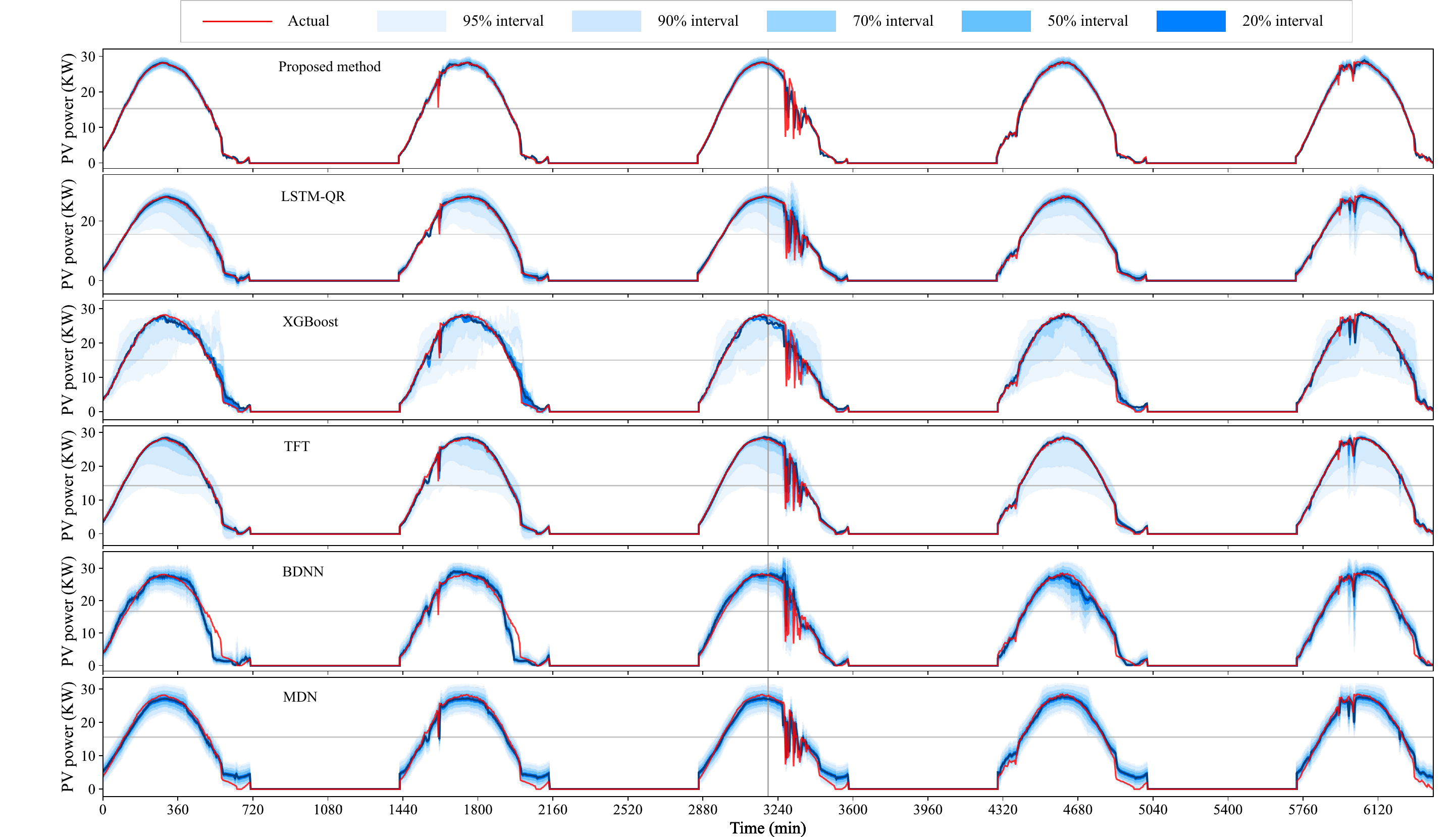}
\caption{Probability prediction results of several methods for dataset \#1}\label{Fig:probability_prediction}
\end{figure}

Fig. \ref{Fig:probability_prediction} shows the prediction curves of the proposed method and benchmark methods on Dataset \#1. The red solid line represents the actual PV output curve, and the blue shaded areas of different intensities represent confidence intervals. It can be clearly seen from the figure that the proposed method has a significant advantage with the smallest error between the predicted median values and actual values. Specifically, this method decomposes and recombines PV power generation into high-frequency and low-frequency components. Based on the characteristics of these two components, different networks are used to extract features from the PV signals, effectively improving prediction accuracy. In the EQN network, confidence width is penalized, effectively enhancing the model's probabilistic prediction capability and making the prediction range smaller. The benchmark methods do not decompose PV power generation but instead use the original power generation data for direct prediction. Whether based on quantile regression methods or neural network methods, their prediction accuracy is not as high as the proposed method. Moreover, since no further penalty is applied to the confidence intervals, their confidence interval widths are significantly larger than those of the proposed method.

\begin{figure}[htbp]
    \centering
    \captionsetup[subfigure]{font=scriptsize,labelfont=scriptsize}
    \subfloat[Proposed method\label{Fig:sub1}]{\includegraphics[width=0.32\linewidth]{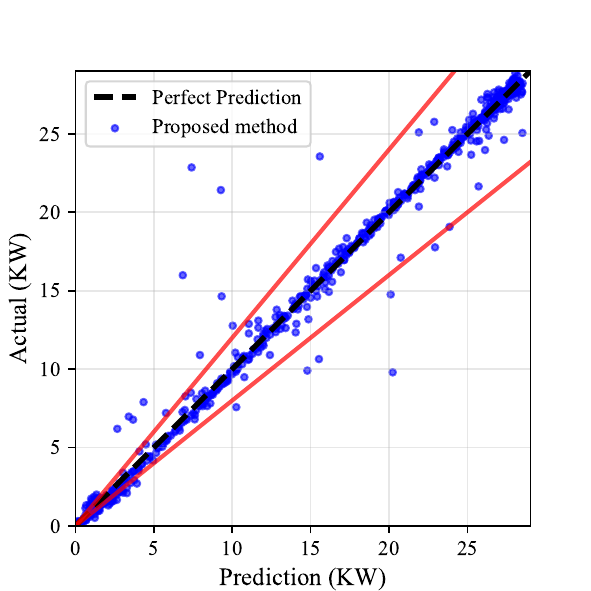}}
    \hfill
    \subfloat[LSTM-QR\label{fig:sub2}]{\includegraphics[width=0.32\linewidth]{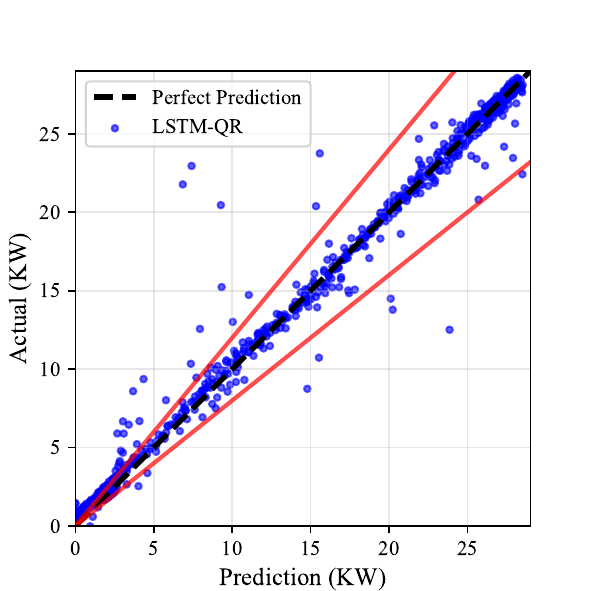}}
    \hfill
    \subfloat[XGBoost\label{fig:sub3}]{\includegraphics[width=0.32\linewidth]{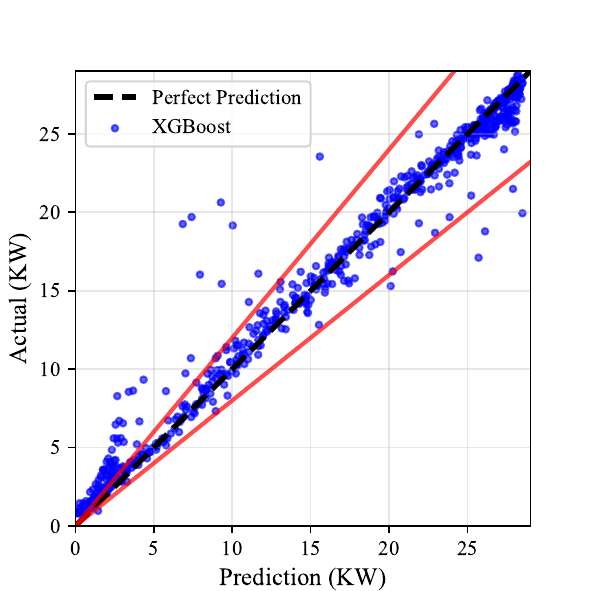}}
    
    
    \subfloat[TFT\label{fig:sub4}]{\includegraphics[width=0.32\linewidth]{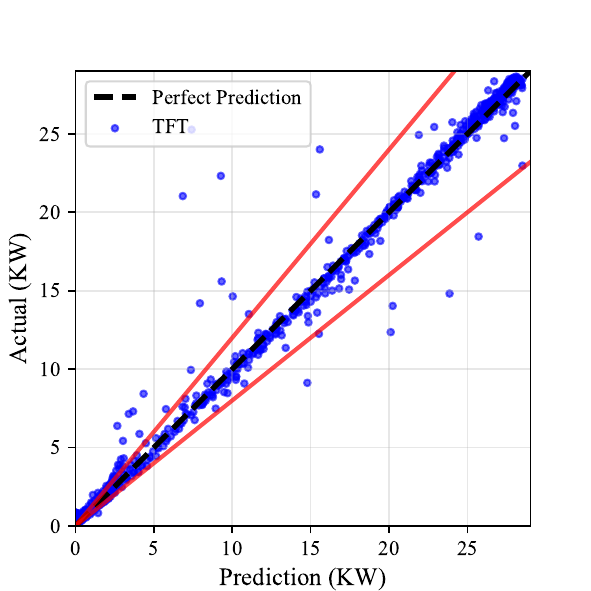}}
    \hfill
    \subfloat[BDNN\label{fig:sub5}]{\includegraphics[width=0.32\linewidth]{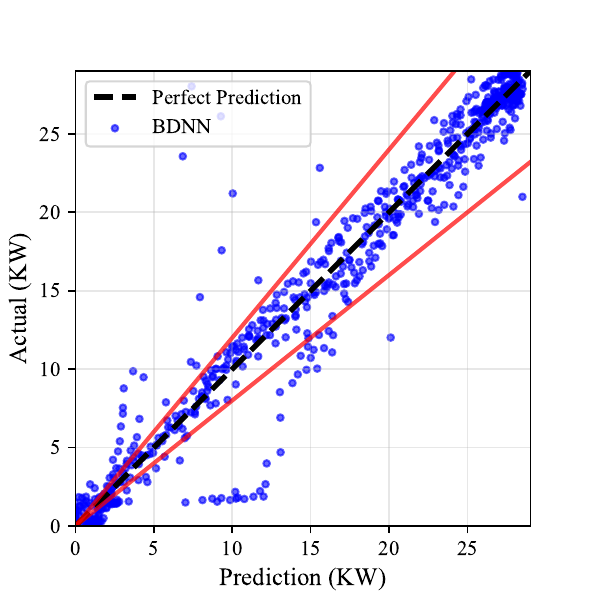}}
    \hfill
    \subfloat[MDN\label{fig:sub6}]{\includegraphics[width=0.32\linewidth]{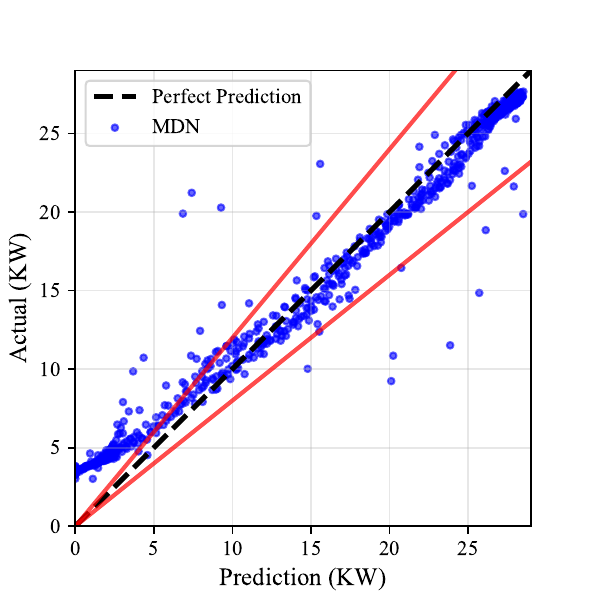}}
    
    \caption{Scatter plot of dataset \#1}
    \label{fig:scatter}
\end{figure}

To further visualize the prediction performance of each model, Fig. \ref{fig:scatter} presents scatter plots comparing actual values and predicted values. In these plots, the x-axis represents predicted values while the y-axis represents actual values. The black solid line represents the ideal prediction line ($y = x$), and the red dashed lines represent the $\pm$20\% confidence interval. From the figure, it can be intuitively seen that the three methods with better prediction performance are the proposed method, LSTM-QR, and TFT. The results of the proposed method show little difference from the actual values throughout the range, and the vast majority of points fall within the confidence interval $\pm$ 20\%. Compared to the proposed method, LSTM-QR and TFT have more points falling outside the ±20\% confidence interval, with LSTM-QR showing the highest degree of dispersion from the ideal prediction line. The three methods XGBoost, BDNN, and MDN perform relatively worse, particularly MDN which shows the greatest deviation from the ideal prediction line in the low power range (0-5 kW), followed by XGBoost, while BDNN has the largest gap from the ideal prediction line within the $\pm$20\% confidence interval. Therefore, we believe that the proposed method has strong prediction capability and will not cause overestimation or underestimation of PV output.
\subsection{Ablation experiment}

We conducted ablation experiments on Dataset \#1. The experimental comparison methods include: 1) The method that directly use IMFs after CEEDMA decomposition for prediction: IMFs-direct; 2) Three methods that use the same network for all high-frequency, low-frequency PV components and meteorological data: CNN-EQN, iTransformer-EQN, and LSTM-EQN; 3) The method No-widthloss without width loss in the EQN network's loss function. Tab. \ref{tab:Ablation} presents the performance metrics of the model ablation experiments.
\begin{table}[htbp]
    \centering
    \caption{Performance metric comparison of different models.}
    \begin{tabular}{p{1.2cm}*{6}{c}} 
    \toprule
    Method&nMAE&nRMSE&$R^2$&CRPS&ACE&WS\\
    \midrule
    Proposed method         &0.0558 &0.1249  &0.9844   &6.1147 &0.1040 &0.8541\\
    IMFs-direct             &0.2551 &0.2551  &0.9349   &10.1836 &0.3805   & 3.1982 \\
    No-widthloss            &0.1388 &0.1724  &0.9703   &27.6131 &0.1963   & 2.5782\\
    CNN-EQN                 &0.2202 &0.2884    &0.9168 &12.7191& 0.4892   &4.7096\\
    iTransformer-EQN        &0.1436  &0.1860    &0.9654 &10.4033 &0.2780   &2.8466\\
    LSTM-EQN                &0.3415  &0.3848   &0.8519&27.8234 &0.4626  &8.0623\\
    \bottomrule
    \end{tabular}
    \label{tab:Ablation}
\end{table}

From the Tab. \ref{tab:Ablation}, it can be observed that the best performing model is the proposed method. Compared to the Proposed method, No-widthloss shows the closest performance on the nMAE and nRMSE metrics, with a deviation of 8.3\% and 4.75\%, respectively. In terms of the $R^2$ metric, compared to the direct IMF and non-width loss, the method proposed in this document achieved an improvement of 4.95\% and 1.41\%. However, in the CRPS and WS metrics, all other methods differ from the proposed method by more than 2, and the ACE metric also shows the proposed method with a significant lead. Therefore, we can conclude that the proposed method, which considers high-frequency and low-frequency components, performs significantly better than directly using IMFs after CEEDMA decomposition for prediction. Adding width constraints to the EQN loss function can improve the confidence of the prediction interval. The approach of using different models to process different data is clearly superior to methods using a single model for processing. This also demonstrates that CNN, iTransformer, and LSTM networks are suitable, respectively, for processing high-frequency data, low-frequency data, and meteorological data.

The computational resource consumption results for each ablation experiment are shown in Tab. \ref{tab:AblationResource}. It can be observed that the proposed method performs optimally both in resource consumption and in training time, with the lowest CPU utilization, the most optimized memory usage, and considerable advantages in TTD, which reflect the efficiency and rationality of the algorithm design. In terms of AET, CNN-EQN has the fastest single-round execution at 0.53 seconds, but its convergence speed is not optimal, and the prediction performance shown in Tab. \ref{tab:Ablation} is also inferior to the Proposed method.
\begin{table}[htbp]
    \centering
    \caption{The algorithm resource consumption results of ablation experiments.}
    \begin{tabular}{p{1.5cm} c c c c c c }
    \toprule
    Method&ACU&AMC& AGMU&AGUR&TTD&AET\\
    \midrule
    Proposed method &1.7\%  &17817.6    &2548.5 &2.8\%  &0.40   & 0.68\\
    IMF-direct&5.7\%  &23306.3    &3536.0 &29.0\% &2.22   & 2.62 \\
    No-widthloss &3.2\%  &27794.8    &4978.4 &17.0\% &2.38   & 1.23 \\
   CNN-EQN         &11.4\% &23425.5    &3302.6 &27.4\% &1.24   &0.53\\
    iTransformer-EQN             &5.8\%  &18064.4    &2074.5 &28.7\% &1.77   &2.05\\
    LSTM-EQN            &6.0\%  &18620.8    &2179.6 &36.8\% &0.48   &0.69\\
    \bottomrule
    \end{tabular}
    \label{tab:AblationResource}
\end{table}

\subsection{Comparisons Among Different Datasets}
To further evaluate the generalization capability of the model, we tested the model's prediction performance on 4 different devices. The raw data for the four devices were collected from July 1, 2022, to September 30, 2022, and the same methods were used for data cleaning and preprocessing. The prediction results were analyzed using evaluation metrics, and the analysis results are shown in Tab. \ref{tab:all results}. In this table, the optimal methods for all indicators in each dataset are highlighted in bold black font for easy reference.
\begin{table*}[htbp]
    \centering
    \caption{Probabilistic forecast result of four datasets.}
    \begin{tabular}{p{2cm} c c c c c c c}
    \toprule
    Method&Datasets&nMAE& nRMSE&$R^2$&CRPS&ACE& WS\\
    \midrule
    \multirow{4}{2em}{Proposed method}
    &\#1&0.0558         &0.1249         &\textbf{0.9844}&\textbf{6.1147}    &0.1040         &\textbf{0.8541}\\
    &\#2&\textbf{0.0744}&0.1577         &\textbf{0.9751}&\textbf{21.6483}   &0.1403         &\textbf{3.3409}\\
    &\#3&0.0631         &0.1386         &\textbf{0.9808}&\textbf{18.5026}   &0.1361         &\textbf{3.0952}\\  
    &\#4&0.0959         &0.1586         &\textbf{0.9748}&\textbf{13.5071}   &0.1792         &\textbf{2.9540}\\
    \cline{2-8}
    \multirow{4}{2em}{LSTM-QR}
    &\#1&0.0735         &0.1464         &0.9786         &19.0842            &0.1212         &1.2509\\
    &\#2&0.2002         &0.2738         &0.9250         &55.3829            &0.0740         &5.5906\\
    &\#3&0.0862         &0.1741         &0.9697         &60.3249            &0.1896         &4.4928\\    
    &\#4&0.1727         &0.2506         &0.9372         &35.9620            &0.0908         &4.8720\\
    \cline{2-8}
    \multirow{4}{2em}{XGBoost}
    &\#1&\textbf{0.0533}&\textbf{0.0983}&0.9745         &22.7904            &0.1641         &1.9733\\
    &\#2&0.0763         &\textbf{0.1227}&0.9562         &88.7254            &0.1262         &7.6124\\
    &\#3&\textbf{0.0459}&\textbf{0.0977}&0.9739         &63.2501            &0.1131         &5.3534\\
    &\#4&\textbf{0.0575}&\textbf{0.1119}&0.9653         &53.2232            &0.0876         &5.8171\\
    \cline{2-8}
    \multirow{4}{2em}{TFT} 
    &\#1&0.0648         &0.1468         &0.9785         &18.2318            &0.1107         &1.0857\\
    &\#2&0.1103         &0.2081         &0.9567         &51.2539            &0.2045         &5.6085\\
    &\#3&0.0902         &0.1874         &0.9649         &50.2684            &0.0868         &4.5294\\
    &\#4&0.1070         &0.1648         &0.9728         &36.8786            &0.0498         &3.1356\\
    \cline{2-8}
    \multirow{4}{2em}{BDNN} 
    &\#1&0.1445         &0.2290         &0.9475         &22.2938            &0.1005         &3.4732\\
    &\#2&0.1559         &0.2193         &0.9519         &50.4663            &0.1115         &11.6504\\
    &\#3&0.2781         &0.3041         &0.9075         &149.9368           &0.0819         &18.5431\\
    &\#4&0.2561         &0.3122         &0.9025         &69.0019            &\textbf{0.0675}& 8.8529\\
    \cline{2-8}
    \multirow{4}{2em}{MDN} 
    &\#1&0.1650         &0.2135         &0.9544         &23.1287            &\textbf{0.0833}&3.1994\\
    &\#2&0.2311         &0.2775         &0.9230         &97.5116            &\textbf{0.0661}&13.8183\\
    &\#3&0.1902         &0.2394         &0.9427         &66.7213            &\textbf{0.0245}&10.5673\\
    &\#4&0.2053         &0.2947         &0.9131         &50.2621            &0.1368         &6.1329\\
    \bottomrule
    \end{tabular}
    \label{tab:all results}
\end{table*}

From the perspective of six performance indicators, the proposed method achieved optimal results on the $R^2$, CRPS, and WS metrics, indicating that the median of the predicted results from the proposed method has the highest degree of fit with the true values. In terms of probabilistic prediction, the proposed method has the smallest deviation between the overall probability distribution and observations, with more accurate quantile intervals, more reliable prediction data, and higher practical value for interval prediction. Regarding the nMAE and nRMSE indicators, except for the proposed method that achieves the best performance on nMAE for Dataset \#2, XGBoost achieved optimal results in all other aspects. This occurs mainly because XGBoost's optimization objective is the MAE for single-point prediction, while the proposed method's optimization objective is to simultaneously optimize 11 quantile predictions and uncertainty estimation, which is primarily reflected in the loss function. The loss function of the proposed method consists of quantile loss, evidence regularization, and confidence interval width constraints, while the XGBoost loss function comprises the objective MAE function and regularization terms. For the ACE indicator, the MDN method achieved optimal results. The main factor contributing to this situation is that MDN directly models based on maximum likelihood estimation methods to simulate the true distribution, simultaneously predicting mean, variance, and mixture weights, outputting a true probability density function. In contrast, the proposed method uses a finite number of quantiles to approximate the complete distribution, and finite quantiles cannot perfectly reconstruct the true distribution. Furthermore, the conversion of evidence to uncertainty can introduce bias, which leads to a poorer performance of the proposed method on the ACE indicator.

\begin{figure}[!htb]
\centering
\includegraphics[width=1.0\columnwidth]{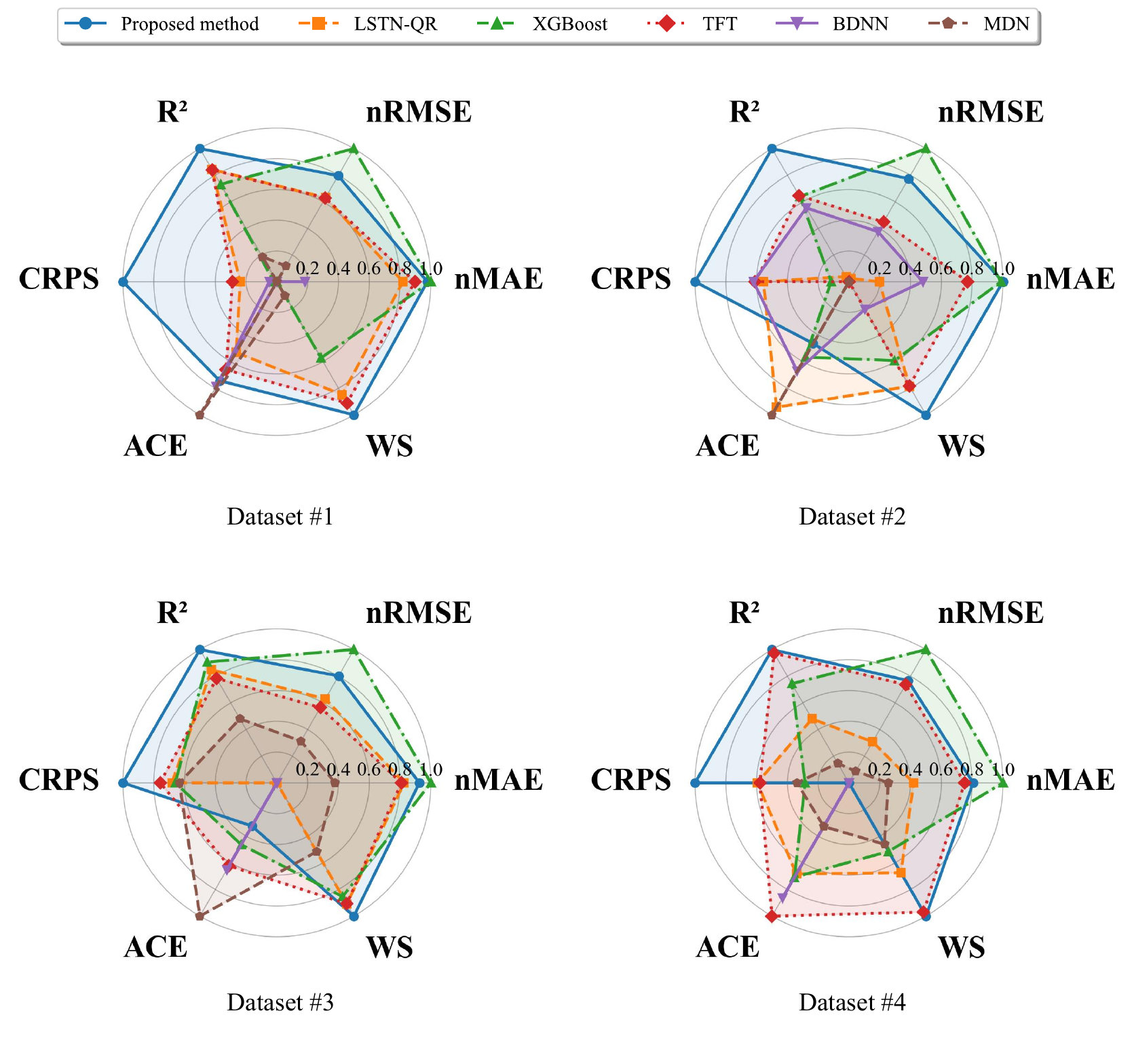}
\caption{Radar Chart of four datasets.}\label{Fig:Radar Chart}
\end{figure}

To more intuitively observe the performance of each method across the four datasets, we plotted radar charts for various indicators of the four datasets, as shown in Fig. \ref{Fig:Radar Chart}. In the charts, the blue solid line with circular points represents the proposed method, the orange dashed line with square points represents LSTM-QR, the green horizontal line with triangular points represents XGBoost, the red dot-dash line with diamond points represents TFT, the purple solid line with inverted triangular points represents BDNN, and the brown dashed line with pentagonal points represents MDN. It can be seen that among the four datasets, all methods achieved the best performance indicator results on Dataset \#1, followed by Datasets \#2 and \#3, while Dataset \#4 showed the worst performance.

In comparison of all methods, the proposed method is the most stable, achieving optimal results for $R^2$, CRPS, and WS, while consistently ranking in the top two for nMAE and nRMSE. Such stability is not possessed by other methods. Combined with Tab. \ref{tab:all results}, taking $R^2$ and WS as examples, the difference between the maximum and minimum values of $R^2$ for the proposed method in the four datasets is less than 0.01, while among the benchmark methods, the maximum difference between maximum and minimum values is 0.0536 (LSTM-QR) and the minimum is 0.0183 (XGBoost). For the WS indicator, the proposed method has a fluctuation value of 2.4868 in the four datasets, while among the benchmark methods, the maximum value is 10.6189 (MDN) and the minimum is 3.6211 (LSTM-QR). Therefore, we believe that the proposed method is the most stable method with the best generalization performance among the six methods.

\begin{figure}[htbp]
    \centering
    \captionsetup[subfigure]{font=scriptsize,labelfont=scriptsize}
    \subfloat[Dataset \#1\label{Fig:boxsub1}]{\includegraphics[width=0.49\linewidth]{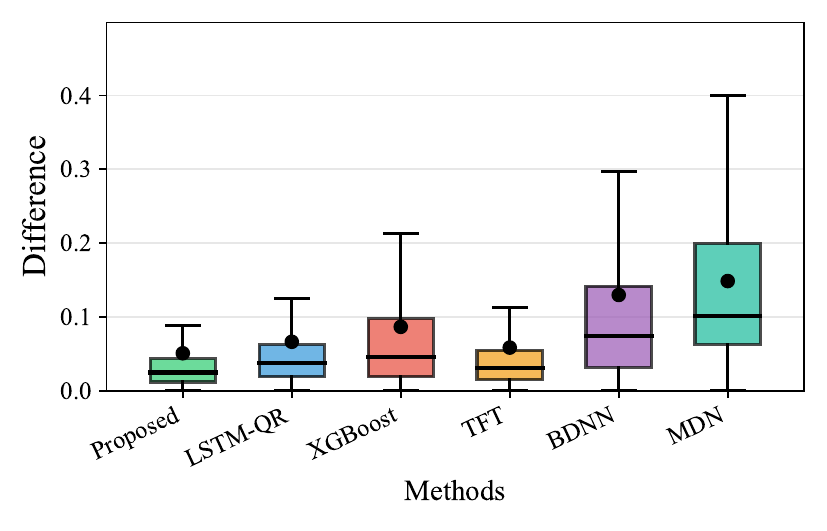}}
    \hfill
    \subfloat[Dataset \#2\label{fig:boxsub2}]{\includegraphics[width=0.49\linewidth]{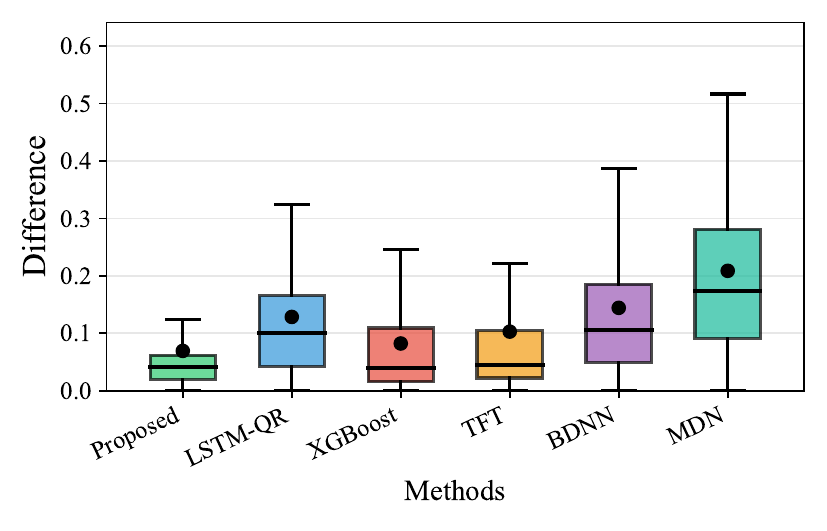}}
    
    
    \subfloat[Dataset \#3\label{fig:boxsub3}]{\includegraphics[width=0.49\linewidth]{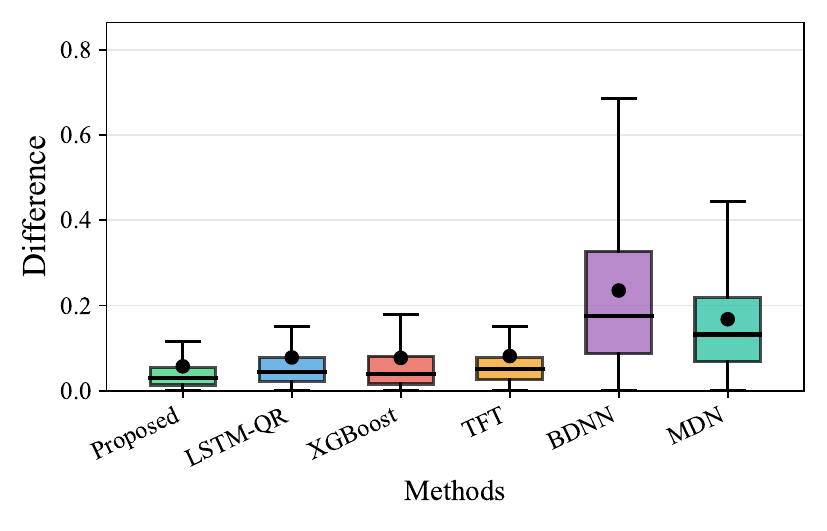}}
    \hfill
    \subfloat[Dataset \#4\label{fig:boxsub4}]{\includegraphics[width=0.49\linewidth]{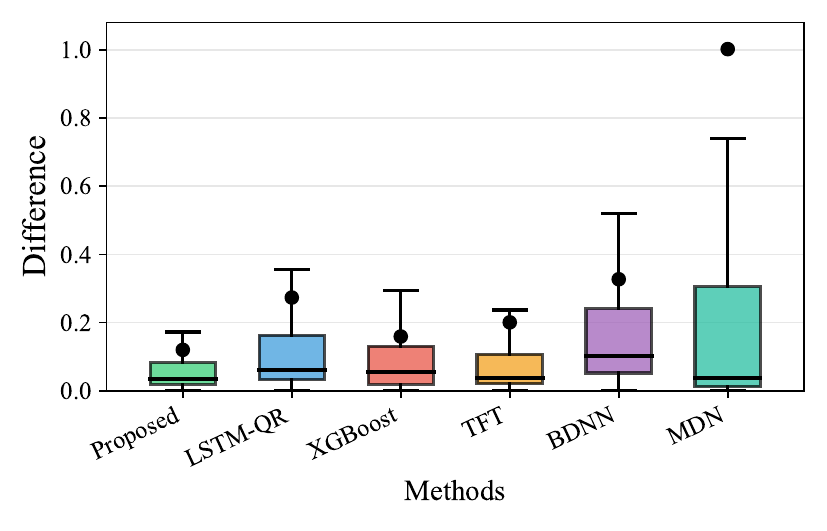}}
    
    \caption{Box plots of four datasets}
    \label{fig:box}
\end{figure}

Furthermore, we present box plots for all data in the four datasets, and the results demonstrate that the proposed method has significant stability and generalization performance, as shown in Fig. \ref{fig:box}. Specifically, across the four datasets, the proposed method exhibits the smallest box width, indicating that the predicted data are more concentrated with lower standard deviation, providing more stable and physically consistent predictions. In contrast, the other algorithms show larger box widths, resulting in weaker generalization performance and lower stability. From the perspective of the median, the proposed method has the lowest median in all four datasets, indicating that this method performs best in terms of error metrics at the middle level, with the smallest prediction error and closer proximity to true values. Among the other methods, LSTM-QR, TFT, and XGBoost perform relatively well; however, their performance across different datasets is not stable. For example, LSTM-QR performs well on Datasets \#1, \#3, and \#4, but poorly on Dataset \#2. The black dots in the box plots represent the mean of the predicted data, used to supplement the display of data central tendency, and are generally analyzed together with the median. If the black dot is higher than the median, it indicates the presence of extremely large values that pull up the average; if the mean is close to the median, it indicates a symmetric data distribution with fewer extreme values and greater stability. Taking Dataset \#2 as an example, the proposed method and LSTM-QR have the smallest deviation between mean and median, indicating that these two methods have a better overall prediction performance. Considering the box width factor, it can be concluded that the proposed method achieves the best results in this regard. The whiskers in the figure represent the data range (minimum and maximum values after excluding outliers). Longer whiskers indicate greater data volatility and the presence of extreme values. Across the four datasets, the proposed method has the shortest whiskers, indicating the highest stability. Therefore, based on the box plots, it can be concluded that the proposed method has superior stability and generalization performance compared to the benchmark methods. This is mainly attributed to decomposing and reintegrating PV power generation into high-frequency and low-frequency components, using appropriate methods for feature extraction, and finally performing feature fusion. This approach can account for both short-term fluctuations and long-term trends, thereby improving overall prediction accuracy, making the prediction results more consistent with actual power generation conditions, enabling the algorithm to adapt to different application scenarios and enhancing model robustness.

\subsection{Comparative Analysis of Algorithm Resource Consumption}
This study conducted model computational resource consumption validation experiments under controlled conditions to evaluate the computational complexity of the model, using Dataset \#1 for the experiments, with results presented in Tab. \ref{tab:Resource}. As shown in Tab. \ref{tab:Resource}, the proposed method achieves an ACU of only 1.7\%, significantly lower than all benchmark methods, indicating that this method has low CPU computational resource requirements and high CPU utilization efficiency. Similarly, the AGMU of 2.8\% is substantially lower than other methods, suggesting a lower algorithmic complexity and more efficient computation. In terms of memory usage, both AMC and AGMU are higher than other algorithms to varying degrees. With respect to computational time, the TTD metric is comparable to TFT, but superior to XGBoost and MDN, while the AET metric only outperforms MDN. In conclusion, the algorithm exhibits the characteristics of trading space for time and computational intensity, making it suitable for environments with sufficient memory but limited computational resources.
\begin{table}[]
    \centering
    \caption{The algorithm resource consumption results of four datasets.}
    \begin{tabular}{p{1.5cm} c c c c c c }
    \toprule
    Method&ACU&AMC& AGMU&AGUR&TTD&AET\\
    \midrule
    Proposed method &1.7\%  &17817.6    &2548.5 &2.8\%  &0.40   & 0.68\\
    LSTM-QR         &6.0\%  &15323.2    &1479.7 &27.3\% &0.34   & 0.36 \\
    XGBoost         &10.4\% &15439.0    &1598.1 &52.2\% &1.00   &$\times$\\
    TFT             &7.3\%  &15413.2    &1500.1 &31.4\% &0.40   &0.60\\
    BDNN            &8.8\%  &15621.1    &1572.7 &25.4\% &0.24   &0.39\\
    MDN             &6.7\%  &15595.8    &1535.7 &36.2\% &2.56   &2.64\\
    \bottomrule
    \end{tabular}
    \label{tab:Resource}
\end{table}
\section{Conclusion}
This paper proposes a CEEDMA-Multi\_nets-EQN method based on CEEDMA decomposition for ultra-short-term prediction of PV power generation. Multiple metrics including deterministic prediction, probabilistic prediction, and time consumption are adopted to evaluate the performance of the model in PV prediction. Ablation experiments are used to validate the performance of each component of the proposed CEEDMA-Multi\_nets-EQN model. Finally, comparisons are made with 5 different baseline models including quantile regression, deep learning networks, and probabilistic network prediction across 4 datasets, evaluating the performance of the proposed method from different aspects. The main conclusions are as follows:
\begin{itemize}
    \item [1.] The IMFs after CEEDMA decomposition are combined into one high-frequency component and one low-frequency component based on frequency thresholds, with CNN and iTransformer used for feature extraction. Experimental data shows that compared to not performing recombination, CEEDMA-Multi\_nets-EQN achieves performance improvements of 1.9\% and 5.52\% on $R^2$, and demonstrates significant reductions in CRPS, ACE, and WS metrics.
    \item[2.] Adding a width penalty term to the loss function of the EQN network improves prediction accuracy by penalizing parts that exceed thresholds. Experimental results show that compared to not adding the width penalty term, nMAE and nRMSE metrics decreased by 8.3]\% and 4.75\% respectively, $R^2$ improved by 1.41\%, and probabilistic prediction metrics also showed significant improvements.
    \item[3.] The proposed method exhibits superior prediction accuracy compared to 5 baseline models on datasets collected from four devices, and demonstrates strong stability. For example, taking WS as an instance, the CEEDMA-Multi\_nets-EQN model shows fluctuations of no more than 2.5 across the four datasets, while the minimum fluctuation of baseline models on this metric is 4.25.
    \item[4.] Computational consumption tests demonstrate the competitive advantages of CEEDMA-Multi\_nets-EQN in actual deployment. For example, the AGMU is 2.8\%, which is significantly lower than other methods, allowing the model to be deployed on devices with lower computational power while achieving good prediction performance.
\end{itemize}

Future work needs to consider computer vision aspects for improving PV prediction accuracy, specifically using cameras to capture sky imagery and adding meteorological information such as solar azimuth angle and cloud cover to the model to further enhance prediction performance.

\bibliographystyle{unsrt}
\bibliography{refer}

\end{document}